\documentclass[lettersize,journal]{IEEEtran}
\usepackage{amsmath,amsfonts}
\usepackage{algorithmic}
\usepackage{algorithm}
\usepackage{array}
\usepackage[caption=false,font=normalsize,labelfont=sf,textfont=sf]{subfig}
\usepackage{textcomp}
\usepackage{stfloats}
\usepackage{url}
\usepackage{verbatim}
\usepackage{graphicx}
\usepackage{cite}
\usepackage{times}
\usepackage{epsfig}
\usepackage{amssymb}
\usepackage{xcolor}
\usepackage{booktabs}
\usepackage{tikz}
\usepackage{multirow}
\usepackage{multicol}
\usepackage{mathtools}
\usepackage{lipsum}
\usepackage{relsize}
\usepackage[T1]{fontenc}
\usepackage{comment}
\usepackage{tikz}
\usepackage{color, colortbl}
\usepackage{hyperref}
\usepackage{booktabs}
\usepackage{makecell}

\newcommand{\methodName}[0]{CoDi}

\definecolor{gold}{HTML}{BD820B}
\definecolor{silver}{HTML}{909090}
\definecolor{bronze}{HTML}{9A5F26}
\definecolor{Gray}{gray}{0.95}

\newcommand{\ie}{\emph{i.e.}}

\newcommand*\circledd[1]{\tikz[baseline=(char.base)]{
            \node[shape=circle,draw,inner sep=0.15pt] (char) {#1};}}      

\newcommand{\first}[1]{%
    {#1\raisebox{0.8pt}{\footnotesize \color{gold} \circledd{1}}}%
}
\newcommand{\second}[1]{%
    {#1\raisebox{0.8pt}{\footnotesize \color{silver} \circledd{2}}}%
}
\newcommand{\third}[1]{%
    {#1\raisebox{0.8pt}{\footnotesize \color{bronze} \circledd{3}}}%
}

\begin{document}

\title{CoDi -- an exemplar-conditioned diffusion model for low-shot counting}

\author{Grega Šuštar, Jer Pelhan, Alan Lukežič, Matej Kristan
  \thanks{All authors are with the Faculty of Computer and Information Science, University of Ljubljana, Slovenia (e-mail: grega.sustar@fri.uni-lj.si; jer.pelhan@fri.uni-lj.si; alan.lukezic@fri.uni-lj.si; matej.kristan@fri.uni-lj.si).}
  \thanks{\textit{Corresponding author: Grega Šuštar}.}
}

\markboth{}%
{Grega Šuštar, Jer Pelhan, Alan Lukežič, Matej Kristan \MakeLowercase{\textit{et al.}}: CoDi -- an exemplar-conditioned diffusion model for low-shot counting}


\maketitle

\begin{abstract}
    Low-shot object counting addresses estimating the number of previously unobserved objects in an image using only few or no annotated test-time exemplars. 
    A considerable challenge for modern low-shot counters are dense regions with small objects. 
    While total counts in such situations are typically well addressed by density-based counters, their usefulness is limited by poor localization capabilities. 
    This is better addressed by point-detection-based counters, which are based on query-based detectors. 
    However, due to limited number of pre-trained queries, they underperform on images with very large numbers of objects, and resort to ad-hoc techniques like upsampling and tiling. 
    We propose \methodName{}, the first latent diffusion-based low-shot counter that produces high-quality density maps on which object locations can be determined by non-maxima suppression.
    Our core contribution is the new exemplar-based conditioning module that extracts and adjusts the object prototypes to the intermediate layers of the denoising network, leading to accurate object location estimation.
    On FSC benchmark, \methodName{} outperforms state-of-the-art by 15\% MAE, 13\% MAE and 10\% MAE in the few-shot, one-shot, and reference-less scenarios, respectively, and sets a new state-of-the-art on MCAC benchmark by outperforming the top method by 38\% MAE. 
    The code is available at \href{https://github.com/gsustar/CoDi}{\textit{https://github.com/gsustar/CoDi}}.
\end{abstract}


\section{Introduction}

\begin{figure}
    \centering
    \includegraphics[width=1\columnwidth]{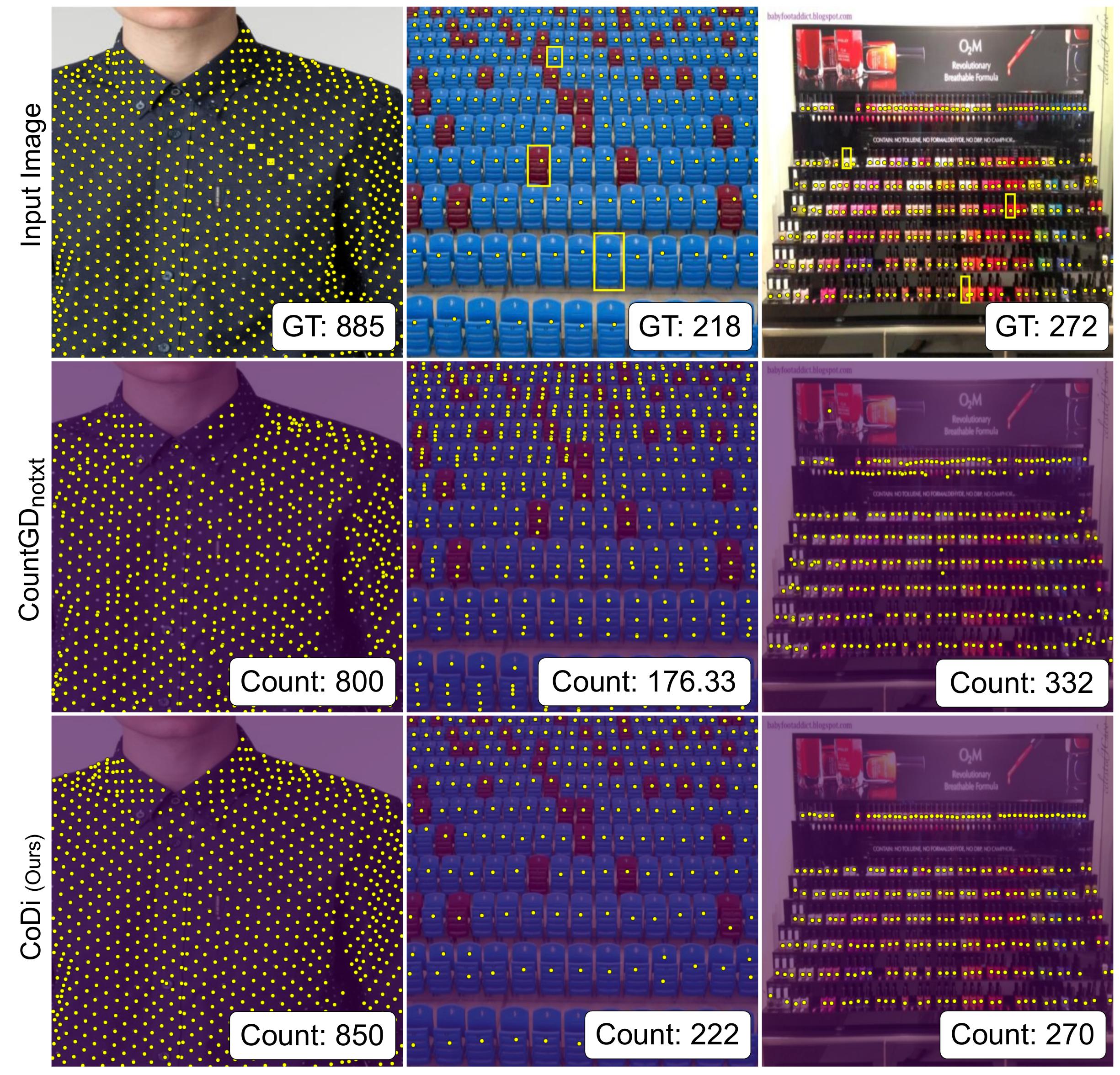}
     \caption{
     State-of-the-art detection-based counters (eg., CountGD~\cite{amini2024countgd}) struggle with small objects in dense regions, and often producing false positives. The proposed \methodName{} employs adaptive prototype extraction and diffusion-based location refinement, leading to accurate detections and counts.
     }
    \label{fig:intro}
\end{figure}

Low-shot object counting estimates the number of objects from unseen categories in an image, given only a few ({\it few-shot / reference-based}) or no exemplars ({\it reference-less}) at test time.
The field has significantly progressed in recent years~\cite{counting-detr, pelhan2024novel, amini2024countgd}, however, state-of-the-art counters still struggle in challenging situations like dense regions with small objects (see Figure~\ref{fig:intro}). 

Dense regions are principally addressed by traditional density-based counters, which estimate the object count by summing a real-valued density map regressed over the input image.
But these approaches are inappropriate for applications, where object locations estimation are required~\cite{zavrtanik2020segmentation,xie2018microscopy}. 
While a non-maxima suppression can be applied to the regressed density map to detect object locations, the regressed densities do not contain well-expressed single modes, in particular on large objects and trigger on the background as well, leading to large number of false positive detections~\cite{counting-detr}.

This weakness is addressed in detection-based counters~\cite{counting-detr,pelhan2024dave, pelhan2024novel, pseco} that predict object locations as bounding boxes and estimate the total counts as the number of detections. 
Since bounding box annotations are required for training, this leads to a much higher annotation challenge compared to density-based counters.
Moreover, detection-based methods underperform in crowded regions, where object centers are separated by only a few pixels, making bounding box predictions unreliable. 

These challenges were circumvented by point-based object-specific counters~\cite{ranasinghe2024crowddiff,liu2023point}, which only predict the object centers instead of bounding boxes.
Such predictions are often sufficient for many practical applications (e.g., cell counting, car counting, production line packing).
Recently, a few-shot point-based counter~\cite{amini2024countgd} was proposed, and demonstrated remarkable performance in class-agnostic counting.
However, since it is formulated as a query-based detector~\cite{liu2024grounding}, its detection and counting capabilities are limited by the number of pre-trained queries.
In dense scenes, the number of objects often exceeds this fixed prediction capacity, thus tiling is applied. 
Images are cut into smaller crops, counted independently, and merged.
This increases computation and introduces a boundary-reconciliation problem, since objects split or duplicated across tile borders are hard to merge into a coherent output.
As a result, the current tiling-based state of the art~\cite{amini2024countgd} only recovers a final scalar count, without a coherent localization output.
Lastly, due to the inherent noise present in the ground truth object centers, i.e. inconsistencies in where the object centers are placed, regression-based models predict the average annotation, which often leads to blurred maps inappropriate for accurate counting by non-maxima suppression. 
As a result, accurate object localization in densely populated regions remains an open challenge.

We address the aforementioned challenges by proposing a new low-shot \underline{Co}unting by \underline{Di}ffusion method (\methodName{}). 
Our main contribution is a new latent diffusion pipeline specialized for low-shot object counting. 
Since diffusion models are trained to predict one instance of annotation, rather than their average, they avoid the problem of annotation noise. \methodName{} exploits this property and converts it into a counting engine capable of producing crisp and detailed object location maps even in highly dense regions of small objects.
Our second contribution is the new exemplar-based conditioning module that injects the annotated exemplars into the denoising network.
The module continuously adapts exemplar prototypes by extracting them during the diffusion process and guiding the denoising network towards the target object locations. 
This leads to output location maps with narrow local kernels appropriate for accurate object localizations through local maxima counting and false positive reduction,
while taking less computational resources compared to the existing pixel-space diffusion counters~\cite{ranasinghe2024crowddiff}. 
As a third contribution, we demonstrate a straight-forward adaptation to a reference-less scenario. To the best of our knowledge, \methodName{} is the first diffusion-based counter that covers the full spectrum of low-shot counting scenarios.

Quantitative evaluation shows that \methodName{} outperforms all current state-of-the-art density-, box- and point-based counters on the challenging FSC147~\cite{ranjan2021learning} few-shot task by 15\% MAE and 29\% RMSE. 
It sets a new state-of-the-art in one-shot scenario, and outperforms the current top-performer~\cite{pelhan2024novel} by 13\% MAE and 49\% RMSE, showcasing remarkable robustness to reducing the number of provided exemplars. 
In the reference-less scenario, \methodName{} outperforms best current state-of-the-art~\cite{pelhan2024novel} by 10\% MAE. 
On the challenging multi-class MCAC~\cite{hobley2025abc} benchmark, \methodName{} sets a new state-of-the-art, outperforming the current top-performer by 38\% MAE and 4\% RMSE.

\section{Related work}
\label{sec:related}

Object counting methods traditionally focused on specific predefined categories such as vehicles \cite{dai2019video}, cells \cite{falk2019u}, people \cite{liu2023point, ranasinghe2024crowddiff, chen2024improving, wang2024efficient, chen2024learning} and polyps \cite{zavrtanik2020segmentation}. These methods required large annotated datasets to train category-specific models, which are often not available. Low-shot counters address this issue by adapting to the object category based on the provided visual exemplars.

\subsection{Density-based methods}

Density-based methods \cite{djukic2023low, hui2025class, liu2022countr, you2023few, shi2022represent, Ranjan_2022_CVPR, finn2017model} approach this challenge by predicting a density map, which when integrated yields the total count. 
CounTR~\cite{liu2022countr} extracts image features via vision transformer and exemplar features with convolutional network and replaces the correlation for template matching with attention operation.
LOCA~\cite{djukic2023low} introduces an object prototype extraction module to construct generalized prototypes with appearance and shape iteratively.
The recently proposed CountDiff~\cite{hui2025class},
adopts an image generation pretrained diffusion model~\cite{rombach2022high} to obtain intermediate attention maps 
that are in the second stage passed through a regression head for density map prediction.
Despite using a diffusion model, 
CountDiff does not rely on a diffusion process to iteratively generate predictions, but rather leverages a strong pretrained backbone as a feature source for density map regression.
YOLO-Count~\cite{zeng2025yolocount} proposes a differentiable zero-shot open-vocabulary counting model that predicts cardinality maps for accurate count estimation and for guiding text-to-image generative models toward better quantity control.
T2ICount~\cite{qian2025t2icount} leverages a pretrained text-to-image diffusion model as a feature extractor and enhances its text sensitivity through a correction module and a novel loss function for better zero-shot counting.
Another weakness of density-based methods is continuous object presence estimation, rather than providing object locations and discrete object counts.

\subsection{Detection-based methods}

Detection-based methods~\cite{counting-detr,pseco,pelhan2024dave,pelhan2024novel, wu2025location} provide explainable outputs in the form of bounding boxes, where the number of detections directly reflects the total count.
PSECO~\cite{pseco} introduces a three-stage point-segment-and-count approach, leveraging SAM~\cite{kirillov2023segment} for segmentation-based proposal generation, followed by a verification step. 
DAVE~\cite{pelhan2024dave} employs a two-stage detect-and-verify strategy, generating high-recall object proposals and refining precision through retrospective verification. 
GeCo~\cite{pelhan2024novel}, enhances SAM with an add-on network, introducing dense object queries to prevent over-generalization and a specialized loss function for dense detection training.
GeCo2~\cite{pelhan2026generalized} is an extension of GeCo, that gradually aggregates
exemplar-specific feature information across scales that leads to more accurate detection of large as well as small objects.
SQLNet~\cite{wu2025location} is a three-stage scale-modulated query and localization network that fully explores the scales of exemplars and achieves effective counting by accurately locating each object and predicting its approximate size.
CountGD++~\cite{amini2026countgdpp} introduces negative exemplars via text and/or visual examples. It further introduces pseudo-exemplars, which automate the annotation of visual examples at inference.
However, detection-based methods still underperform in crowded scenes and require expensive full bounding box annotations during training to achieve state-of-the-art counting results.

\subsection{Localization-based methods}

Localization-based methods provide discrete counts by estimating object center points, which has proven effective in crowded scenes for object-specific counting~\cite{ranasinghe2024crowddiff, chen2022task}.
S2MC2~\cite{chen2022task} is a self-supervised multi-category counter specialized for a practical application that leverages point-level supervision to lower the labeling cost and to make predictions in a class incremental setting.
CountSE~\cite{liu2025countse} is a text-guided zero-shot object counting algorithm that
generates multiple precise soft exemplars at different scales to enhance counting models driven solely by semantics.
The recently proposed CountGD~\cite{amini2024countgd} is an open-world counter built on top of GroundingDINO~\cite{liu2024grounding}, leveraging both text and visual prompts to predict object center points. However, object detectors operate with a limited number of queries, necessitating tiled inference and additional post-processing.
VQCounter~\cite{ye2025vqcounter} extends CountGD~\cite{amini2024countgd} with a visual prompt queue that stores detected bounding boxes during training to increase exemplar diversity, together with a Voronoi-diagram-based matching cost that better exploits point-level annotations. Since it inherits the query-based GroundingDINO detector backbone, it remains subject to the same fixed-query capacity limits as CountGD.

CrowdDiff~\cite{ranasinghe2024crowddiff} is a class-specific pixel-space diffusion counter, specialized for crowd counting.
The method progressively generates narrow-kernel density maps using the reverse diffusion process~\cite{song2019generative, ho2020denoising}.
While \methodName{} and CrowdDiff do share a similar idea of location map generation, they differ substantially in their task formulation, conditioning strategy, and output representation.
Additionally, during inference~\cite{ranasinghe2024crowddiff}
combines multiple predictions to improve the counting performance and optionally performs tiling of the input image, thus requiring massive amounts of computational resources and significantly slowing down the inference process.

In contrast with the existing localization-based counters, \methodName{} is not limited with the maximum number of queries inherent to transformer-based architectures like CountGD~\cite{amini2024countgd}.
Additionally, \methodName{} performs diffusion in latent space and does not require tiling of input images which makes it computationally efficient compared to related methods, such as CountGD and CrowdDiff.
Lastly, \methodName{} formulates the core object detection mechanism through a diffusion process. The denoising process is trained to directly generate narrow, separated location-map peaks, unlike CountDiff~\cite{hui2025class}, which uses a frozen pretrained text-to-image diffusion model as a feature source for density regression.

\section{CoDi -- Counting with diffusion}  \label{sec:CoDi}

\begin{figure}
    \centering
    \includegraphics[width=\linewidth]{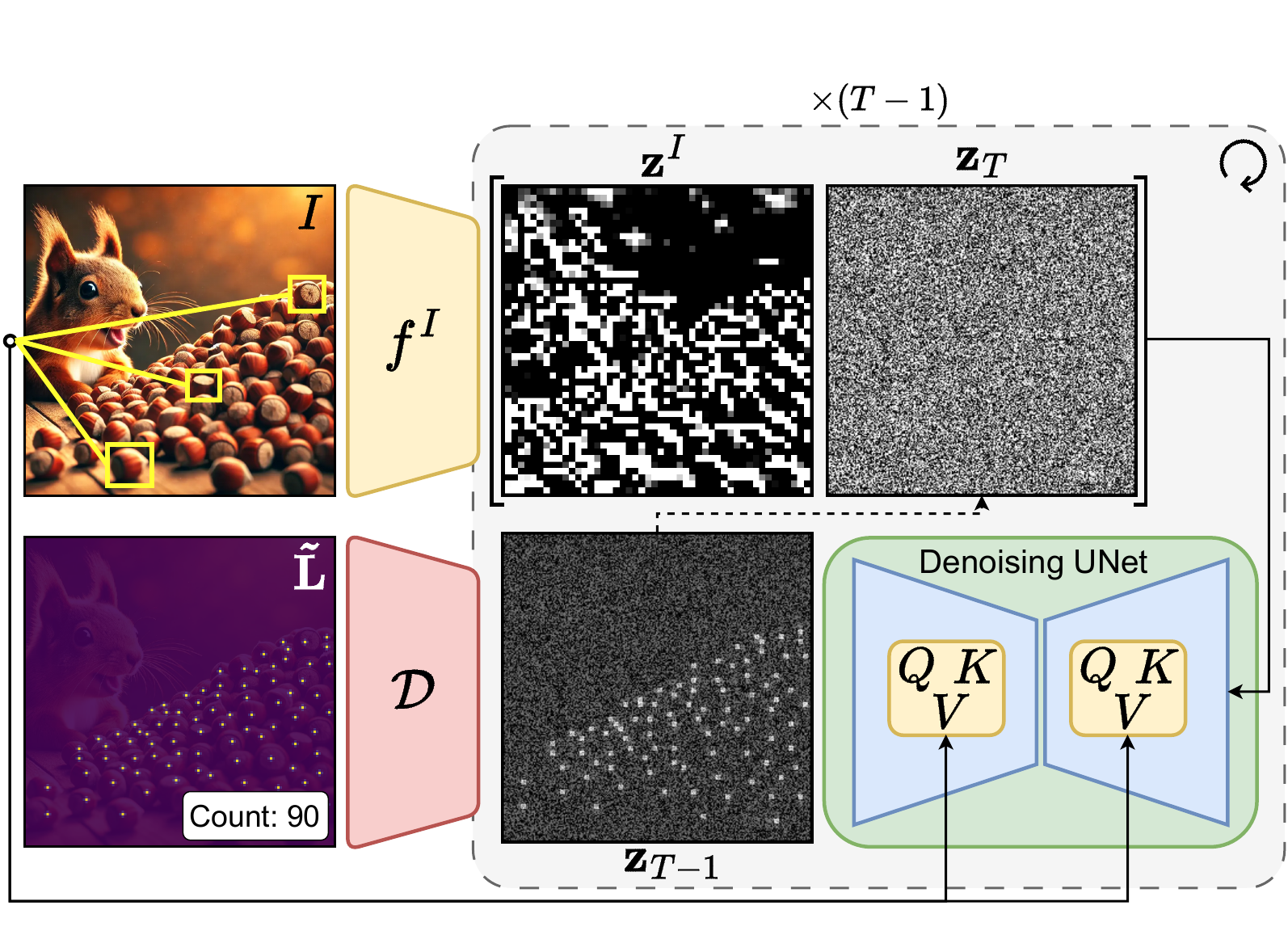}
    \caption{\methodName{} employs a latent conditioned diffusion model with adaptive prototype extraction and a modulated Denoising UNet to gradually refine object locations map $\tilde{\mathbf{L}}$.}
    \label{fig:architecture}
\end{figure}

Given an input image $I \in \mathbb{R}^{H_0 \times W_0 \times 3}$ and a set of $K_E$ exemplar bounding boxes $\boldsymbol{B}^E = \{b_i\}_{i=1:K_E}$, the task is to estimate the number $N$ of all objects of the specified category. 
We approach this problem from detection perspective, by inferring an object locations map $\mathbf{L} \in \mathbb{R}^{H_0 \times W_0}$ with each object represented by a normalized Gaussian, from which 2D object center locations $\mathbb{L} = \{ (x_j, y_j) \}_{j=1:N}$ can be trivially extracted (e.g., using non-maxima suppression).
The proposed Counting-with-Diffusion counter (CoDi) is a localization low-shot object counting method, with architecture outlined in Figure~\ref{fig:architecture}.

We apply a conditioned latent diffusion model (LDM)~\cite{rombach2022high} as the core framework that gradually refines the latent object locations map. For the forward diffusion simulation (i.e., in training), the object locations map $\mathbf{L} \in \mathbb{R}^{H_0 \times W_0 \times 1}$ is constructed by placing a normalized Gaussian with a small variance $\sigma^2$ over each ground truth object location. The locations map is then projected by a pre-trained  variational auto-encoder (VAE) into the latent representation $\mathbf{z}_0 = \mathcal{E}(\mathbf{L})$, and transformed into a sequence of increasingly noisy/diffused representations $\mathbf{z}_t$ governed by the following conditional pdf
\begin{equation}
    q(\mathbf{z}_t|\mathbf{z}_0) = \mathcal{N}(\mathbf{z}_t; \sqrt{\bar{\alpha}_t} \mathbf{z}_0, (1-\bar{\alpha}_t) \mathbf{I}), \label{eq:marginal}
\end{equation}
where $t \in \{0,...,T\}$ and $\bar{\alpha}_t \coloneqq \prod_{s=0}^{t} (1 - \beta_s)$, with $\beta_s$ a well-behaved noise schedule that defines the Gaussian noise variance at diffusion step $t$.

An inverse diffusion network $f^{\epsilon}(\mathbf{z}_t, t, \mathbf{y}_t)$ is then trained to reverse the step $t$, using a step-varying conditioning variable $\mathbf{y}_t$.
The network is implemented as a standard three-level UNet~\cite{ronneberger2015u}. 
The number of channels in the first level is set to $c_0 = 192$, with each subsequent level doubling the number of channels ($c_i = 2c_{i-1}$). 
Each level of the network is composed of two residual blocks, with an additional down/upsampling block when transitioning between levels.

\subsection{Counting-specific diffusion model}  \label{sec:counting_diffusion}
\begin{figure}
    \centering
    \includegraphics[width=0.8\linewidth]{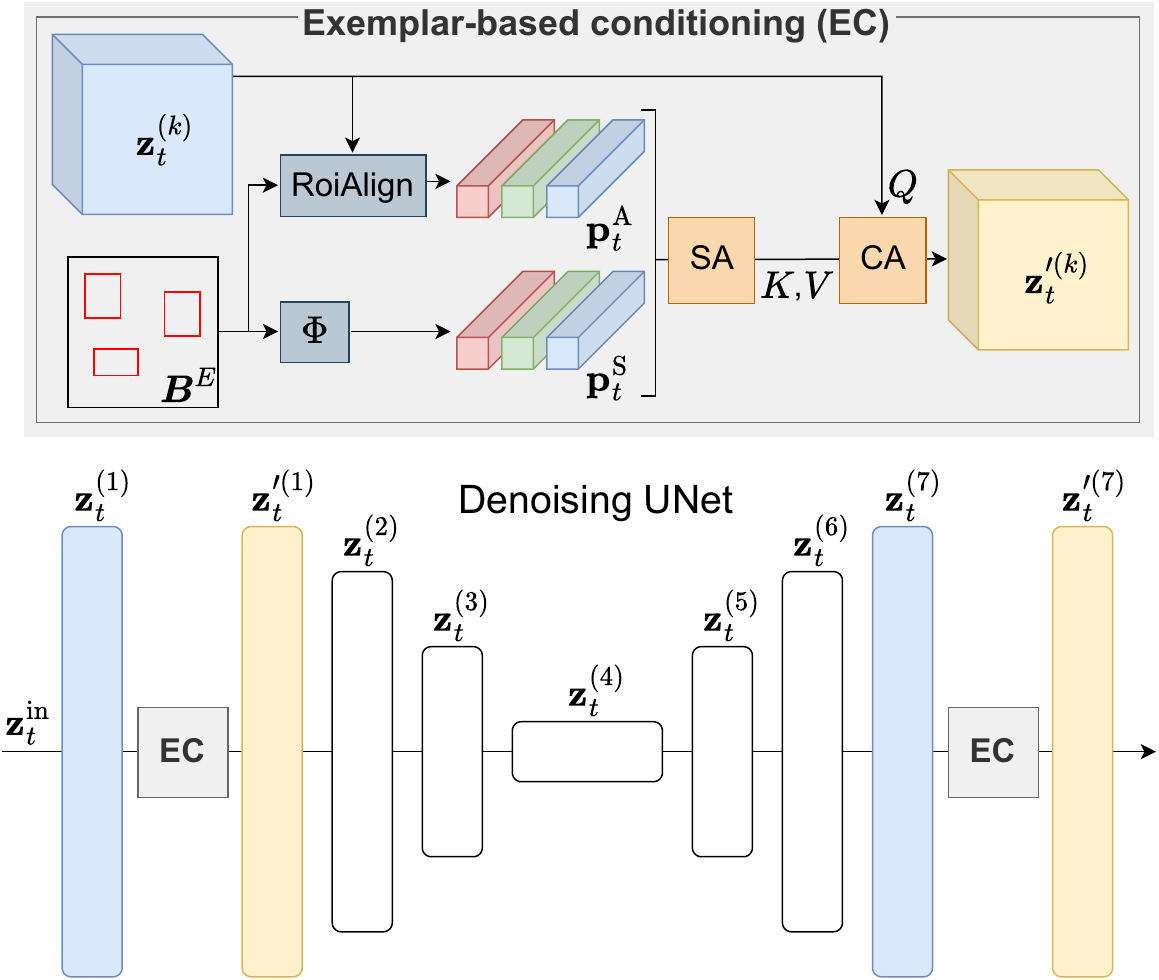}
    \caption{The exemplar-based conditioning module (EC) is applied in the first and last layer of the Denoising UNet.}
    \label{fig:conditioning}

\end{figure}

In the low-shot counting scenarios, the inverse diffusion network has to specialize to the specific object category selected by the input exemplars and ignore other objects in the image. We ensure this test-time specialization by \textit{exemplar-based conditioning} described in the following.
To enable access to the image content,
the input RGB image $I \in \mathbb{R}^{H_0 \times W_0 \times 3}$ is encoded into $\mathbf{z}^I \in \mathbb{R}^{h \times w \times d'}$ using a pre-trained 
RADIO backbone~\cite{ranzinger2024radio}, i.e., $\mathbf{z}^I = f^I(I)$. The latent image is then concatenated with the current latent locations map $\mathbf{z}_t$
along the channel dimensions, i.e., $\mathbf{z}_t^\text{in}=\text{cat}(\mathbf{z}^I,\mathbf{z}_t)$, forming the input to the denoising UNet. 

The proposed exemplar-based conditioning then
modulates the individual layers of the UNet using the locations of user-specified exemplars (Figure~\ref{fig:conditioning}). 
Since the features change with diffusion steps, the modulation is designed to be adaptive as well.
At timestep $t$, layer-specific prototypes $\mathbf{p}_t^{(k)}$ are constructed from the UNet $k$-th layer features $\mathbf{z}_t^{(k)}$ as follows.
An appearance prototype $\mathbf{p}_{t,i}^\text{A} \in \mathbb{R}^{1 \times d_k}$ corresponding to the $i$-th exemplar bounding box $b_i$ is constructed by RoI-aligning~\cite{he2017mask} features from $\mathbf{z}_t^{(k)}$ into a $1\times d_k$ and transforming them by a linear projection $\mathbf{W}_k$, i.e.,
\begin{equation}
\mathbf{p}_{t,i}^\text{A} = \mathbf{W}_k \cdot\text{RoIAlign}(\mathbf{z}_t^{(k)}, b_i).
\end{equation}
Following the standard practice~\cite{djukic2023low}, a shape prototype $\mathbf{p}_{t,i}^\text{S} \in \mathbb{R}^{1 \times d_k}$ is extracted as 
\begin{equation}
\mathbf{p}_{t,i}^\text{S} = \Phi([{b_i^W, b_i^H}]),
\end{equation}
where $\Phi(\cdot)$ is a two-layer MLP~\cite{djukic2023low}, while $b_i^W$ and $b_i^H$ are $i$-th exemplar width and height, respectively. 
The $k$-th layer prototypes $\mathbf{p}_t^{(k)} \in \mathbb{R}^{2K_E \times d_k}$ are thus obtained by concatenating all appearance and shape prototypes and transforming them by a self-attention operation.
 
The exemplar-based conditioning (Figure \ref{fig:conditioning}) is then implemented as a single transformer layer injected right after the UNet $k$-th layer. 
The transformer uses a multi-head attention with three heads and with 
$\mathbf{z}_t^{(k)}$ acting as queries and the layer-specific prototypes $\mathbf{p}_t^{(k)}$ as keys and values. 
The exemplar-based conditioning layers are injected only to the first and last UNet layers, where the resolution is still sufficiently high to enable small objects detection.

\subsection{Reference-less adaptation}  \label{sec:zero-shot}

In a reference-less setup, exemplars are not provided and the task is to count the majority-class objects. Due to the lack of exemplars, the prototypes for the exemplar-based conditioning cannot be extracted directly. To compensate for this, we introduce layer-specific trainable reference-less queries $\mathbf{q}^{(k)} \in \mathbb{R}^{K_Z \times d_k}$, which are converted to layer-specific prototypes
$\mathbf{p}_t^{(k)} \in \mathbb{R}^{K_Z \times d_k}$ by attending them to the features $\mathbf{z}_t^{(k)}$ in the $k$-th layer. 
CoDi is thus easily adapted to the reference-less scenario by taking CoDi trained for the few-shot counting, freezing all parameters and training only the reference-less queries and the corresponding cross-attention with the layer features.

\subsection{Training objective}  \label{subsec:training_objective}

CoDi is trained using the v-parametrization~\cite{salimans2022progressive, lin2024common} which tasks the inverse diffusion network to predict $\mathbf{v}_t \coloneqq \sqrt{\bar\alpha_t}\epsilon - \sqrt{1 - \bar\alpha_t}\mathbf{z}_0$, and derives required equations to move from $\mathbf{z}_t$ into $\mathbf{z}_{t-1}$. 
The training objective is thus
\begin{equation}  \label{eq:L_with_v}
    L_{\text{v}} = \mathbb{E}_{t, \mathbf{z}_0, \mathbf{y}, \epsilon \sim \mathcal{N}(0, \textbf{I})}[\left\Vert \mathbf{v}_t - \tilde{\mathbf{v}}_t\right\Vert^2_2], 
\end{equation}
where $\tilde{\mathbf{v}}_t$ is predicted by the inverse diffusion UNet, i.e., $\tilde{\mathbf{v}}_t = f^{\epsilon}(\mathbf{z}_t, t, \mathbf{y}_t)$.

An additional variational lower bound term (VLB)~\cite{nichol2021improved} is added to $L_{\text{v}}$, resulting in the final hybrid training objective
\begin{equation}  \label{eq:L_final}
    L  = L_{\text{v}} + \lambda_{\text{vlb}} L_{\text{vlb}}, 
\end{equation}
where $\lambda_{\text{vlb}}$ is a weight of the VLB term.

\subsection{Low-shot counting application}  \label{subsec:inference}

At application time, only the input image is provided in a reference-less setup, and few user-annotated exemplars in a few-shot setup.
The input to the denoising network is thus a concatenation of the conditioning latent input image, i.e., $\mathbf{z}^I = f^I(I)$, and the latent object locations map initialized by a random sample from the isotropic Gaussian, i.e., $\mathbf{z}_T \sim \mathcal{N}(0,\mathbf{I})$. 
The layer-specific prototypes are constructed at each iteration of the denoising network, which delivers an estimate of the 
final denoised latent object locations map $\tilde{\mathbf{z}}_0$ after $T$ iterations. The latent map is then decoded by the pretrained VAE decoder~\cite{podell2023sdxl} into the image-resolution object locations map 
$\tilde{\mathbf{L}} \in \mathbb{R}^{H_0 \times W_0 \times 1}$. Object centers are detected as local maxima in $3 \times 3$ neighborhoods exceeding a small threshold $\tau_\text{max} =0.1$. 
The total object count is then estimated as the number of detected maxima, adjusted by the standard  test-time normalization~\cite{liu2022countr, amini2024countgd}.

\section{Experiments}
\label{sec:exp}

\subsection{Implementation details}  \label{subsec:imp_details}

The input images are encoded by the AM-RADIOv2.5-L~\cite{ranzinger2024radio} backbone with a patch size 16 to derive the input-level conditioning for the denoising UNet from Section~\ref{sec:counting_diffusion}.
To enrich the embedding, 
the intermediate features from levels [5, 11, 17, 23]
are concatenated and transformed by $1\times1$ convolution to reduce the number of channels to $d=160$.

For encoding ($\mathcal{E}$) and decoding ($\mathcal{D}$) the latent object location map embeddings we re-use the RGB image VAE~\cite{podell2023sdxl} pretrained for image-reconstruction
with four-channel latent space, as is standard practice in the image generation field.
To match the encoder input requirements in training time, the 2D object locations map is expanded into three identical channels. 
For the same reason, the output latent object locations map decoded by VAE is a three-channel tensor. The final decoded object locations map is thus obtained by averaging the three output channels.
 
During inference we run 25 diffusion steps with the default DDPM sampler and a uniform stride from~\cite{nichol2021improved}.
Following established test-time practices~\cite{pelhan2024dave, pelhan2024novel, counting-detr, shi2022represent}, 
the input image is upsampled to $W_0 = H_0 = 1024$ when the average area of the exemplars is less than 1250 pixels.

\subsection{Training details} 
\label{subsec:train}

All training images are resized to $W_0 = H_0 = 512$, before applying standard augmentations such as random horizontal flipping and tiling \cite{liu2022countr}.
The size of the normalized Gaussians in the ground truth locations maps are set to $\sigma = 0.5$.

\methodName{} is trained from scratch using the hybrid loss described in (\ref{eq:L_final}), with $\lambda_{\text{vlb}} = 0.001$. 
We use $T = 1000$ timesteps and a scaled linear noise scheduler~\cite{rombach2022high}. 
Additionally, we enforce a zero-terminal SNR (Signal-to-Noise Ratio) following~\cite{lin2024common} for consistent behavior during training and inference at timestep at $t=T$.
\methodName{} is trained for 150 epochs using AdamW optimizer and a warmup cosine scheduler with a peak learning rate of $1e^{-4}$ after 5 epochs. 
Both gradient clipping of 1.0 and weight decay of 0.0001 are used. 
We also employ an EMA rate of 0.9999, to prevent excessive overfitting. 
The number of exemplars for the few-shot setup is kept constant, as in other related works and set to $K_E=3$. 
\methodName{} is trained on a single H100 GPU with a batch size 16, which takes about 15 hours. 

\subsection{Evaluation metrics and dataset}
\label{subsec:eval_and_dataset}

\methodName{} is evaluated on two challenging datasets: FSC147 \cite{ranjan2021learning} and  MCAC \cite{hobley2025abc}. The FSC147 \cite{ranjan2021learning} dataset consists of 6135 images and 147 different object classes. 
The dataset is split into 3659 training, 1286 validation and 1190 test images. 
The object categories between splits are mutually exclusive, \ie, the target object categories in the validation and test splits are not observed during the training phase. 
Each image is annotated with three bounding boxes referencing the exemplars and point annotations corresponding to each object in the image that is of the same class as the exemplars. 

MCAC \cite{hobley2025abc} is a recently proposed counting dataset tailored for class-agnostic and multi-class counting. 
It consists of 4756 training images with 287 classes, 2413 validation images with 37 classes, and 2114 testing images with 19 classes.
Similarly to FSC147 the classes in different splits are disjoint. Each image may contain between 1 and 4 different object classes and between 1 and 300 objects per class. 
During training, we take exemplar bounding boxes randomly from instances with less than 30\% occlusion. 
Following the established evaluation protocol, we use bounding boxes of the three least occluded instances and evaluate performance on multi-class images separately for each present class. Following standard practice, the occlusion limit is set to 70\% for all three splits.
The counting accuracy is evaluated by the standard performance measures~\cite{ranjan2021learning, you2023few, shi2022represent}, i.e., Mean Absolute Error (MAE) and Root Mean Square Error (RMSE).

\subsection{State-of-the-art comparison}  \label{subsec:res}

\textbf{Few-shot counting.} 
\methodName{} is compared with the following state-of-the-art density-based counters, which provide global count estimation: CountDiff~\cite{hui2025class}, CACViT~\cite{wang2024vision}, LOCA~\cite{djukic2023low}, CounTR~\cite{liu2022countr}, SafeCount~\cite{you2023few}, VCN~\cite{Ranjan_2022_CVPR}, BMNet+~\cite{shi2022represent}, CFOCNet~\cite{yang2021class}, FamNet~\cite{ranjan2021learning}, MAML~\cite{finn2017model}, GMN~\cite{lu2019class}. 
In addition, the detection-based counters
GeCo~\cite{pelhan2024novel}, GeCo2~\cite{pelhan2026generalized}, CountGD++$_{\text{notxt}}$~\cite{amini2026countgdpp}, DAVE~\cite{pelhan2024dave}, PSECO~\cite{pseco}, SAM-C~\cite{samcount}, C-DETR~\cite{counting-detr}
are also included in the comparison. 
Finally, we include CountGD~\cite{amini2024countgd}, the most recent localization-based counter, which similarly as \methodName{}, provides object locations by center points.
To ensure a fair comparison, all existing methods were evaluated using the inference configurations recommended by the authors, including upsampling, tiling, and cropping techniques when applicable.

\begin{table}
    \centering
    \caption{
    Few-shot performance on FSC147~\cite{ranjan2021learning} of density-based (top), detection-based (middle) and localization-based (bottom) methods. $(\cdot)^\dagger$ indicates that the method also uses text prompts.
    }
    \label{tab:fsc147-results}
    \resizebox{\linewidth}{!}{
    \begin{tabular}{lllll}
        \toprule
        & \multicolumn{2}{c}{Validation set} & \multicolumn{2}{c}{Test set} \\  
        \cmidrule(lr){2-3} \cmidrule(lr){4-5} 
        Method & MAE ($\downarrow$) & RMSE($\downarrow$) & MAE($\downarrow$) & RMSE($\downarrow$) \\ 
        
        \midrule
        GMN~\cite{lu2019class}{\smaller[3] ACCV18}           & 29.66 & 89.81 & 26.52 & 124.57 \\
        MAML~\cite{finn2017model}{\smaller[3] ICML17}        & 25.54 & 79.44 & 24.90 & 112.68 \\
        FamNet~\cite{ranjan2021learning}{\smaller[3] CVPR21} & 23.75 & 69.07 & 22.08 & 99.54  \\
        CFOCNet~\cite{yang2021class}{\smaller[3] WACV21}     & 21.19 & 61.41 & 22.10 & 112.71 \\
        BMNet+~\cite{shi2022represent}{\smaller[3] CVPR22}      & 15.74 & 58.53 & 14.62 & 91.83  \\
        VCN~\cite{Ranjan_2022_CVPR}{\smaller[3] CVPRW22}     & 19.38 & 60.15 & 18.17 & 95.60  \\
        SAFEC~\cite{you2023few}{\smaller[3] WACV23}          & 15.28 & 47.20 & 14.32 & 85.54  \\
        CounTR~\cite{liu2022countr}{\smaller[3] BMVC22}      & 13.13 & 49.83 & 11.95 & 91.23  \\
        LOCA~\cite{djukic2023low}{\smaller[3]ICCV23}         & 10.24 & 32.56 \third{} & 10.79 & 56.97  \\
        CACViT~\cite{wang2024vision}{\smaller[3]AAAI24}      & 9.13  & 48.96 & 10.63 & 37.95 \third{} \\
        CountDiff$^\dagger$~\cite{hui2025class}{\smaller[3] ECCV2024}  & 8.43  & 31.03 \second{} & 9.24  & 53.41 \\
        \midrule
        
        C-DETR~\cite{counting-detr}{\smaller[3] ECCV22}      & 20.38 & 82.45  & 16.79 & 123.56 \\
        SAM-C~\cite{samcount}{\smaller[3] arXiv23}           & 31.20 & 100.83 & 27.97 & 131.24 \\
        PSECO~\cite{pseco}{\smaller[3] CVPR24}               & 15.31 & 68.36  & 13.05 & 112.86 \\
        DAVE~\cite{pelhan2024dave}{\smaller[3] CVPR24}       & 9.75  & 40.30  & 10.45 & 74.51  \\
        GeCo~\cite{pelhan2024novel}{\smaller[3] NeurIPS24}   & 9.52  & 43.00  & 7.91  & 54.28  \\
        GeCo2~\cite{pelhan2026generalized}{\smaller[3] AAAI26} & 9.40  & 33.28  & 7.64\third{} & 39.39  \\
        CountGD++$_{\text{notxt}}$~\cite{amini2026countgdpp}{\smaller[3] CVPR26} & -- & -- & 8.10 & 35.40 \second{} \\
        \midrule
        CountGD$_{\text{notxt}}$~\cite{amini2024countgd}{\smaller[3] NeurIPS24}    & 8.12 \second{} & 38.97 & 8.35 & 89.80 \\
        CountGD$^\dagger$~\cite{amini2024countgd}{\smaller[3] NeurIPS24}           & \textbf{7.10} \first{} & \textbf{26.08} \first{} & 6.75 \second{} & 43.65 \\
        \methodName{} {\smaller[3](Ours)}                                         & 8.21 \third{} & 35.67 & \textbf{5.74} \first{} & \textbf{30.99} \first{}  \\
        
        \bottomrule
    \end{tabular}
    }
\end{table}


Counting performance on the FSC147 dataset is presented in Table~\ref{tab:fsc147-results}. 
\methodName{} outperforms the current state-of-the-art few-shot density-based counter CountDiff~\cite{hui2025class} by 38\% MAE and 42\% RMSE on the test set.
Among the methods grouped as detection-based counters, \methodName{} outperforms the strongest MAE competitor GeCo2~\cite{pelhan2026generalized} by 25\% MAE and the strongest RMSE competitor CountGD++$_{\text{notxt}}$~\cite{amini2026countgdpp} by 12\% RMSE on the test set.
Similarly, \methodName{} outperforms the recent localization-based counter CountGD~\cite{amini2024countgd} by 15\% MAE and 29\% RMSE on the test set, using only information from three exemplars.
CountGD offers two inference settings: one using both visual exemplars and text prompts, and CountGD$_{\text{notxt}}$, using only visual exemplars. Since CoDi does not use text prompts, CountGD$_{\text{notxt}}$ is the prompt-matched comparison, while the original CountGD (with text) is reported as a stronger multimodal reference.
Compared to the version without text prompts (CountGD$_{\text{notxt}}$), that uses only visual exemplars, \methodName{} outperforms it by even a larger margin, i.e., 31\% MAE and 65\% RMSE.

\begin{table}[h!]
    \centering
    \caption{
    Localization performance of CoDi and CountGD$_{\text{notxt}}$ on the FSC147 dataset
    }
    \begin{tabular}{lllllll}
        \toprule
        & \multicolumn{3}{c}{Validation set} & \multicolumn{3}{c}{Test set} \\  
        \cmidrule(lr){2-4} \cmidrule(lr){5-7}
         Method & R ($\uparrow$) & P ($\uparrow$) & F1 ($\uparrow$) & R ($\uparrow$) & P ($\uparrow$) & F1 ($\uparrow$) \\
        
        \midrule
        
        CountGD$_{\text{notxt}}$ & 0.85 & 0.88 & 0.86 & 0.81 & 0.95 & 0.87 \\ 
        \methodName{} & 0.93  & 0.88 & 0.90 & 0.88 & 0.95 & 0.92  \\
        \bottomrule
        \label{tab:localization}
    \end{tabular}
\end{table}

Following a point-based detection protocol~\cite{zavrtanik2020segmentation}, we evaluate the localization quality of \methodName{} and compare it to the current state-of-the-art CountGD$_{\text{notxt}}$.
By assigning each predicted center point to the closest ground-truth center and checking if it lies within the bounding box of the object TP, FP and FN are obtained. 
We then compute precision (P), recall (R) and F1 score on the FSC147 dataset.
The results are presented in Table~\ref{tab:localization}. 
\methodName{} substantially outperforms CountGD$_{\text{notxt}}$ with nearly $\sim$8\% points better recall, while maintaining the same precision. 
The average distance of TP detections to ground-truth centers is 6.89 pixels on the validation set and 8.79 pixels on the test set.
These results show \methodName's superior detection capability compared to the competing localization-based counter and provide additional insights to the localization performance.

We further evaluate \methodName{} on MCAC~\cite{hobley2025abc} to demonstrate its generalization capabilities. 
Table~\ref{tab:mcac-results} shows that \methodName{} outperforms the best exemplar-based method GeCo2~\cite{pelhan2026generalized} by 38\% MAE and 4\% RMSE, 
and the best exemplar-free method ABC123~\cite{hobley2025abc} by 44\% MAE and 7\% RMSE on the test set. 
This experiment further supports the results from Table~\ref{tab:fsc147-results} and demonstrates CoDi's strong ability to distinguish between objects from different categories present in an image, given only a few exemplars.

\begin{table}[ht!]
    \centering
    \caption{
       Few-shot performance on MCAC \cite{hobley2025abc}. The exemplar-free ABC123~\cite{hobley2025abc} is added for a complete comparison. 
    }
    \label{tab:mcac-results}
    \resizebox{\linewidth}{!}{
    \begin{tabular}{lllll}
        \toprule
        & \multicolumn{2}{c}{Validation set} & \multicolumn{2}{c}{Test set} \\  
        \cmidrule(lr){2-3} \cmidrule(lr){4-5} 
        Method & MAE ($\downarrow$) & RMSE($\downarrow$) & MAE($\downarrow$) & RMSE($\downarrow$) \\ 
        
        \midrule
        
        ABC123\cite{hobley2025abc}{\smaller[3] ECCV25}        & 8.96 \second{} & 15.93 \second{} & 9.52 \third{}  & 17.64 \third{} \\
        
        \midrule
        
        FamNet \cite{ranjan2021learning}{\smaller[3] CVPR21}  & 24.76 & 41.12 & 26.40 & 45.52 \\ 
        BMNet+ \cite{shi2022represent}{\smaller[3] CVPR22}    & 15.83 & 27.07 & 17.29 & 29.83 \\ 
        GeCo \cite{pelhan2024novel}{\smaller[3] NeurIPS24}    & 15.31 & 28.42 & 19.91 & 33.25
        \\
        CounTR \cite{liu2022countr}{\smaller[3] BMVC22}       & 15.07 & 26.26 & 16.12 & 29.28 \\
        LOCA \cite{djukic2023low}{\smaller[3] ICCV23}         & 10.45 & 20.81  & 10.91 & 22.04
        \\
        GeCo2 \cite{pelhan2026generalized}{\smaller[3] AAAI26} & 9.38  \third{} & 18.80 \third{} & 7.93 \second{} & 17.05 \second{}
        \\
        \methodName{} {\smaller[3] (Ours)}                    & \textbf{5.81} \first{}   & \textbf{15.26} \first{} & \textbf{5.26} \first{} & \textbf{16.31} \first{} \\
    
        \bottomrule
    \end{tabular}
    }
\end{table}

\textbf{One-shot counting.} 
A one-shot counting setup, i.e., counting with a single exemplar, is considered to further explore exemplar-based counting robustness.
Results on FSC147~\cite{ranjan2021learning} are reported in Table~\ref{tab:fsc147-1shot-results}. 
Note that \methodName{} does not require any specialization (e.g., additional re-training or fine-tuning) for one-shot counting, 
yet outperforms all state-of-the-art one-shot density- and detection-based methods. 
In particular, it outperforms GeCo by 13\% MAE and 49\% RMSE on the test set.
Compared to the three-shot setup, the one-shot \methodName{} achieves even slightly lower RMSE but does not drastically increase MAE. 
This result indicates similar counting capability on both setups, which demonstrates great robustness to the number of input exemplars. 

\begin{table}[ht!]
    \centering
    \caption{
    One-shot performance on FSC147~\cite{ranjan2021learning} of density-based (top), detection-based (middle) and localization-based (bottom) methods. $(\cdot)^\dagger$ indicates that the method also uses text prompts.
    }
    \label{tab:fsc147-1shot-results}
    \resizebox{\linewidth}{!}{
    \begin{tabular}{lllll}
        \toprule
        & \multicolumn{2}{c}{Validation set} & \multicolumn{2}{c}{Test set} \\  
        \cmidrule(lr){2-3} \cmidrule(lr){4-5} 
        Method & MAE ($\downarrow$) & RMSE($\downarrow$) & MAE($\downarrow$) & RMSE($\downarrow$) \\ 

        \midrule

        GMN~\cite{lu2019class}{\smaller[3] ACCV18}                    & 29.66 & 89.81 & 26.52 & 124.57 \\
        CFOCNet~\cite{yang2021class}{\smaller[3] WACV21}              & 27.82 & 71.99 & 28.60 & 123.96 \\
        FamNet~\cite{ranjan2021learning}{\smaller[3] CVPR21}          & 26.55 & 77.01 & 26.76 & 110.95 \\
        BMNet+~\cite{shi2022represent}{\smaller[3] CVPR22}               & 17.89 & 61.12 & 16.89 & 96.65  \\
        CounTR~\cite{liu2022countr}{\smaller[3] BMVC22}               & 13.15 & 49.72 & 12.06 & 90.01  \\
        LOCA$_\text{1-shot}$~\cite{djukic2023low}{\smaller[3] ICCV23} & 11.36 & 38.04 & 12.53 & 75.32  \\
        CountDiff$^\dagger$~\cite{hui2025class}{\smaller[3]ECCV2024}            & 9.35 \second{}  & \textbf{33.44} \first{} & 10.29 \third{} & 57.27 \second{}  \\

        \midrule
        
        PSECO~\cite{pseco}{\smaller[3] CVPR24}                         & 18.31 & 80.73 & 14.86 & 118.64 \\
        DAVE$_\text{1-shot}$~\cite{pelhan2024dave}{\smaller[3] CVPR24} & 10.98 & 43.26 & 11.54 & 86.62  \\
        GeCo~\cite{pelhan2024novel}{\smaller[3] NeurIPS24}             & 9.97 \third{}  & 37.85 \third{} & 8.10 \second{}  & 60.16 \third{}  \\

        \midrule
			
        \methodName{} {\smaller[3] (Ours)}                               & \textbf{8.92} \first{}  & 35.89 \second{} & \textbf{7.04} \first{}  & \textbf{30.23} \first{} \\
        
        \bottomrule
    \end{tabular}
    }
\end{table}

\textbf{Reference-less counting.}
To cover the whole low-shot counting spectrum, the reference-less performance on the FSC147 dataset is reported in Table~\ref{tab:0shotfsc147-results}.  
In this setup, exemplars are not provided, and counters are required to count the majority object class. 
As described in Section~\ref{sec:zero-shot}, minimal adaptation of the few-shot \methodName{} is required to support this setup, with the number of reference-less queries set to $K_Z = 3$.
Table~\ref{tab:0shotfsc147-results} shows that \methodName{} achieves the best test MAE among methods that do not use text prompts. 
Compared to the best recent text-free state-of-the-art GeCo~\cite{pelhan2024novel}, \methodName{} delivers slightly better RMSE and outperforms the method by 10\% MAE.
We also include PSECO~\cite{pseco}, CountDiff~\cite{hui2025class} and YOLO-Count~\cite{zeng2025yolocount} which use text prompts only
and can be thus considered as {\it zero-shot} methods. 
\methodName{} outperforms these counters by 28\% MAE, 19\% MAE and 23.6\% respectively.
While \methodName{} stays competitive when compared to the recently proposed text-prompted counter T2ICount~\cite{qian2025t2icount}, CountSE~\cite{liu2025countse}, a novel text-prompted zero-shot method, achieves stronger overall results.
Based on these results we conclude that \methodName{} successfully covers the whole low-shot counting spectrum and remains competitive without relying on text prompts.

\begin{table}
    \centering
    \caption{
    Reference-less and zero-shot counting on FSC147~\cite{ranjan2021learning} for density-based (top), detection-based (middle) and localization-based (bottom) methods. $(\cdot)^\dagger$ indicates that the method uses text prompts (i.e., zero-shot setup)}. 
    \label{tab:0shotfsc147-results}
    \resizebox{\linewidth}{!}{
    \begin{tabular}{lllll}
        \toprule
        & \multicolumn{2}{c}{Validation set} & \multicolumn{2}{c}{Test set} \\  
        \cmidrule(lr){2-3} \cmidrule(lr){4-5} 
        Method & MAE ($\downarrow$) & RMSE($\downarrow$) & MAE($\downarrow$) & RMSE($\downarrow$) \\ 
        \midrule
        RepRPN-C~\cite{ranjan2022exemplar}{\smaller[3] ACCV22} & 29.24 & 98.11 & 26.66 & 129.11 \\
        RCC~\cite{hobley2022learning}{\smaller[3] arXiv22}     & 17.49 & 58.81 & 17.12 & 104.5  \\
        CounTR~\cite{liu2022countr}{\smaller[3] BMVC22}        & 17.40 & 70.33 & 14.12 & 108.01 \\
        LOCA~\cite{djukic2023low}{\smaller[3] ICCV23}          & 17.43 & 54.96 \third{} & 16.22 & 103.96 \\
        CountDiff$^\dagger$~\cite{hui2025class}{\smaller[3] ECCV2024} & 15.50 & \textbf{54.33} \first{} & 14.83 & 103.15 \\
        T2ICount$^\dagger$~\cite{qian2025t2icount}{\smaller[3] CVPR25} & 13.78 \second{} & 58.78 & 11.76 \second{} & 97.86 \third{} \\
        
        \midrule
        
        PSECO$^\dagger$~\cite{pseco}{\smaller[3] CVPR24}       & 23.90 & 100.33 & 16.58 & 129.77 \\
        DAVE~\cite{pelhan2024dave}{\smaller[3] CVPR24}         & 15.71 & 60.34  & 15.51 & 116.54 \\
        YOLO-Count$^\dagger$~\cite{zeng2025yolocount}{\smaller[3] ICCV25} & 15.43 & 58.36 & 14.80 & 96.14 \second{} \\
        GeCo~\cite{pelhan2024novel}{\smaller[3] NeurIPS24}     & 14.81 & 64.95  & 13.30 & 108.72 \\
        \midrule
        CountSE$^\dagger$~\cite{liu2025countse}{\smaller[3] CVPR25} & \textbf{8.51} \first{} & 54.93 \second{} & \textbf{7.84} \first{} & \textbf{82.99} \first{} \\
        \methodName{}{\smaller[3] (Ours)}                      & 14.14 \third{} & 64.87 & 11.97 \third{} & 106.00 \\
         
        \bottomrule
    \end{tabular}
    }
\end{table}

\subsection{Computational requirements}


\begin{table}[htbp]
\centering
\footnotesize
\caption{Computational requirements benchmark on the FSC-147 validation set on A100 40GB GPU.}
\label{tab:benchmark}
\setlength{\tabcolsep}{4pt}

\begin{tabular}{l rrrr r}
\toprule
& \multicolumn{4}{c}{Wall-clock time (s / img)} & \\
\cmidrule(lr){2-5}
Method & Mean & Std & Min & Max & FPS \\
\midrule
LOCA    & 0.036 & 0.002 & 0.034 & 0.089  & 27.65 \\
CountGD & 1.058 & 3.433 & 0.539 & 89.232 & 0.95  \\
GeCo    & 0.675 & 0.496 & 0.451 & 6.840  & 1.48  \\
CoDi    & 1.177 & 0.269 & 0.982 & 1.760  & 0.85  \\
\bottomrule
\end{tabular}

\vspace{4pt}

\begin{tabular}{l rrr r r}
\toprule
& \multicolumn{3}{c}{VRAM (GB)} & & \\
\cmidrule(lr){2-4}
Method  & Mean & Min  & Max   & \makecell{FLOPs\\(GFLOPs)} & \makecell{Params\\(M)} \\
\midrule
LOCA    & 1.48 & 1.48 & 1.48  & 80.1   & 36.9  \\
CountGD & 7.43 & 7.07 & 19.55 & 1255.7 & 232.8 \\
GeCo    & 9.41 & 8.69 & 21.63 & 3331.3 & 646.4 \\
CoDi    & 8.47 & 6.90 & 13.0  & 2528.1 & 196.0 \\
\bottomrule
\end{tabular}
\end{table}

Computation costs of \methodName{} are benchmarked on the FSC147 dataset and compared to CountGD~\cite{amini2024countgd}, a localization-based counter, LOCA~\cite{djukic2023low}, a density-based counter, and GeCo~\cite{pelhan2024novel}, a detection-based counter. 
Table~\ref{tab:benchmark} reports wall-clock time, VRAM usage, FLOPs, and parameter count on an NVIDIA A100 40GB GPU. As expected, the density-based LOCA is by far the fastest and lightest method. Among the remaining methods, \methodName{} shows the most stable inference behavior, with a mean wall-clock time of 1.1s, which is comparable to CountGD's 1.0s and slightly higher than GeCo's 0.67s. 
However, \methodName{}'s variance and worst-case latency are drastically lower than CountGD's, whose runtime degrades sharply on images with many small objects. 
This is because \methodName{}'s computation cost only depends on the input image resolution and has a fixed number of forward passes during inference. 
CountGD, on the other hand, adaptively tiles the image based on the size of the exemplars which leads to a rapid increase in the number of forward passes, when the objects in the image are small.
\methodName{} also requires less VRAM than GeCo and has the fewest parameters among the non-density-based methods.
The results also show that FLOPs do not correlate well with inference time. 
This is because modern GPUs execute simple, highly parallelizable floating-point operations very efficiently, while more complex, higher-level operations can cause disproportionate slowdowns regardless of raw FLOP count. 
For instance, operations such as CountGD's tiling, weigh more heavily on wall-clock time than raw FLOP count would suggest.

\subsection{Ablation study}  \label{subsec:ablation}

\begin{figure}
    \centering
    \includegraphics[width=\linewidth]{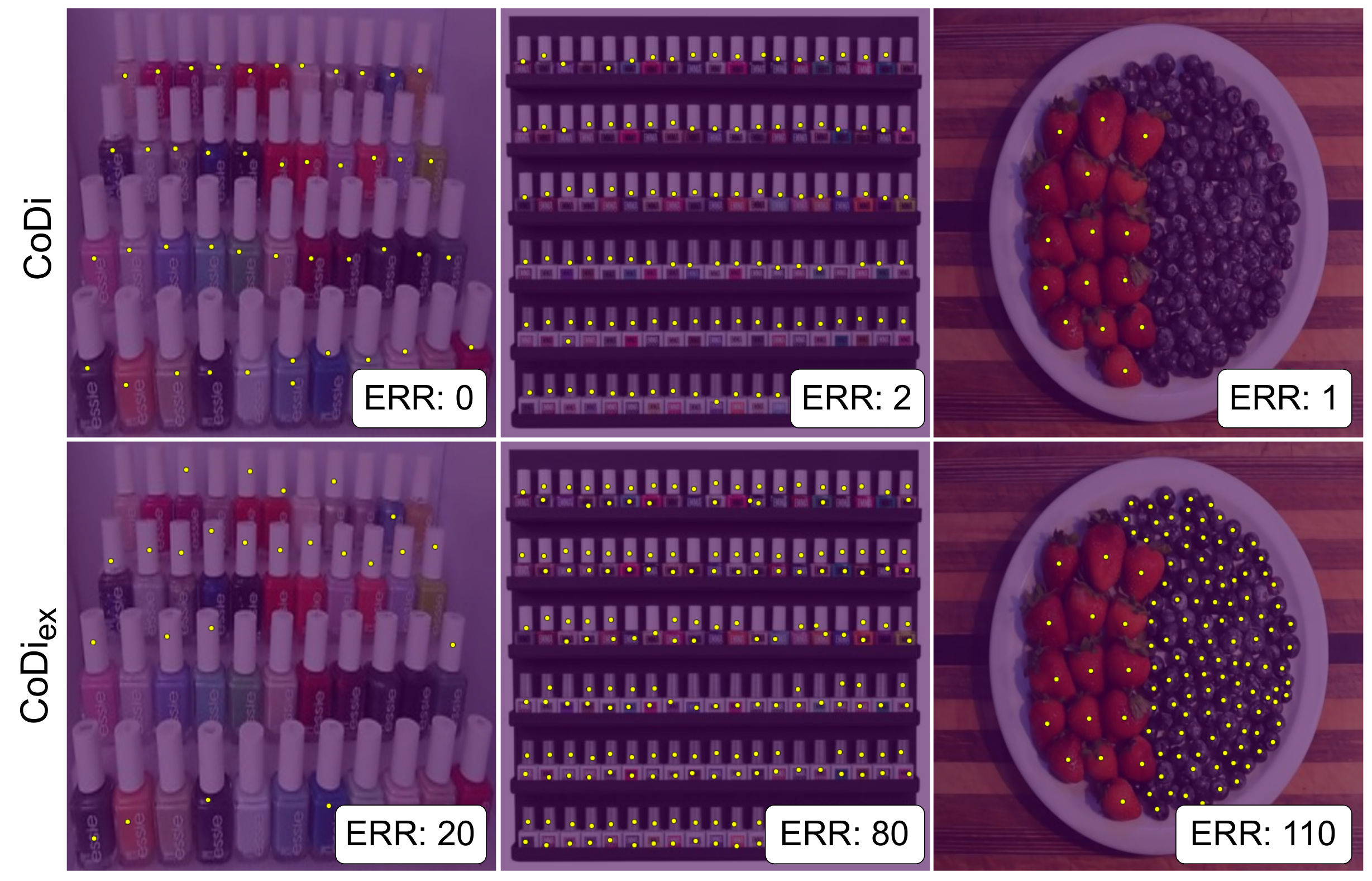}
    \caption{
    \methodName{} shows better discrimination between different categories and robustness to double detections and false negatives compared to the version with external conditioning (\methodName{}$_{\text{ex}}$). Yellow dots denote detections, while ERR is count error. 
    }
    \label{fig:ex_vs_in_quali}
\end{figure}

We next perform an ablation study to analyze the parameters and architectural designs.
Results are reported in Table~\ref{tab:ablation_fsc147}.
First, we verify the importance of the proposed exemplar-based conditioning in the denoising UNet by replacing it with a traditional external conditioning from~\cite{rombach2022high}, denoted by \methodName{}$_{\text{ex}}$.  
In particular, prototypes in \methodName{}$_{\text{ex}}$ are constructed only once, before the denoising network and remain unchanged during all $T$ denoising iterations. This is done by RoI pooling from the image features.
A substantial performance drop of 14\% MAE and 13\% RMSE is observed, compared to the proposed \methodName{}. 
We provide an additional qualitative comparison of \methodName{} and \methodName{}$_{\text{ex}}$ in Figure \ref{fig:ex_vs_in_quali}, demonstrating the importance of the proposed exemplar based conditioning for object category discrimination and robustness to double and missed detections.
We further test whether this gain can be attributed simply to any form of temporal exemplar adaptivity, rather than to conditioning on features aligned with the same timestep and layer.
We construct \methodName{}$_{\text{mod}}$, which modulates prototypes extracted once before denoising with learnable timestep and layer embeddings similar to~\cite{peebles2023scalable}.
\methodName{}$_{\text{mod}}$ achieves 8.46/38.44 MAE/RMSE on the validation set, an improvement over \methodName{}$_{\text{ex}}$, further verifying that temporal adaptivity matters.
However, a performance drop of 3\%/7.7\% relative to \methodName{} can still be observed, showing that it is not only the temporal information that matters, but also more importantly using the features from the same timestep and layer.

\begin{table}
    \centering
    \caption{
    Ablation study on the FSC147~\cite{ranjan2021learning} validation set.
    }
    \label{tab:ablation_fsc147}
    \resizebox{0.7\columnwidth}{!}{
    \begin{tabular}{lll}
        \toprule
         Method & MAE ($\downarrow$) & RMSE($\downarrow$) \\
        
        \midrule
        
        \methodName{}                                 & \textbf{8.21}  & \textbf{35.67} \\
        \methodName{}$_{\text{ex}}$                   & 9.62 & 40.92 \\
        \methodName{}$_{\text{mod}}$         & 8.46 & 38.44 \\
        \methodName{}$_{\overline{\mathrm{p}^s}}$     & 10.81 & 52.76 \\
        \methodName{}$_{\text{VAE}}$                  & 11.78 & 42.38 \\
        \methodName{}$_{\text{DINOv2}}$               & 9.67 & 40.18 \\
        \methodName{}$_{\text{ResNet}}$      & 10.42 & 39.89 \\
        \methodName{}$_{\overline{TTN}}$              & 8.20 & 36.34 \\
        \methodName{}$_{\text{direct}}$ & 15.42 & 43.79 \\

        \bottomrule
    \end{tabular}
    }
\end{table}

\begin{figure}
    \centering
    \includegraphics[width=\linewidth]{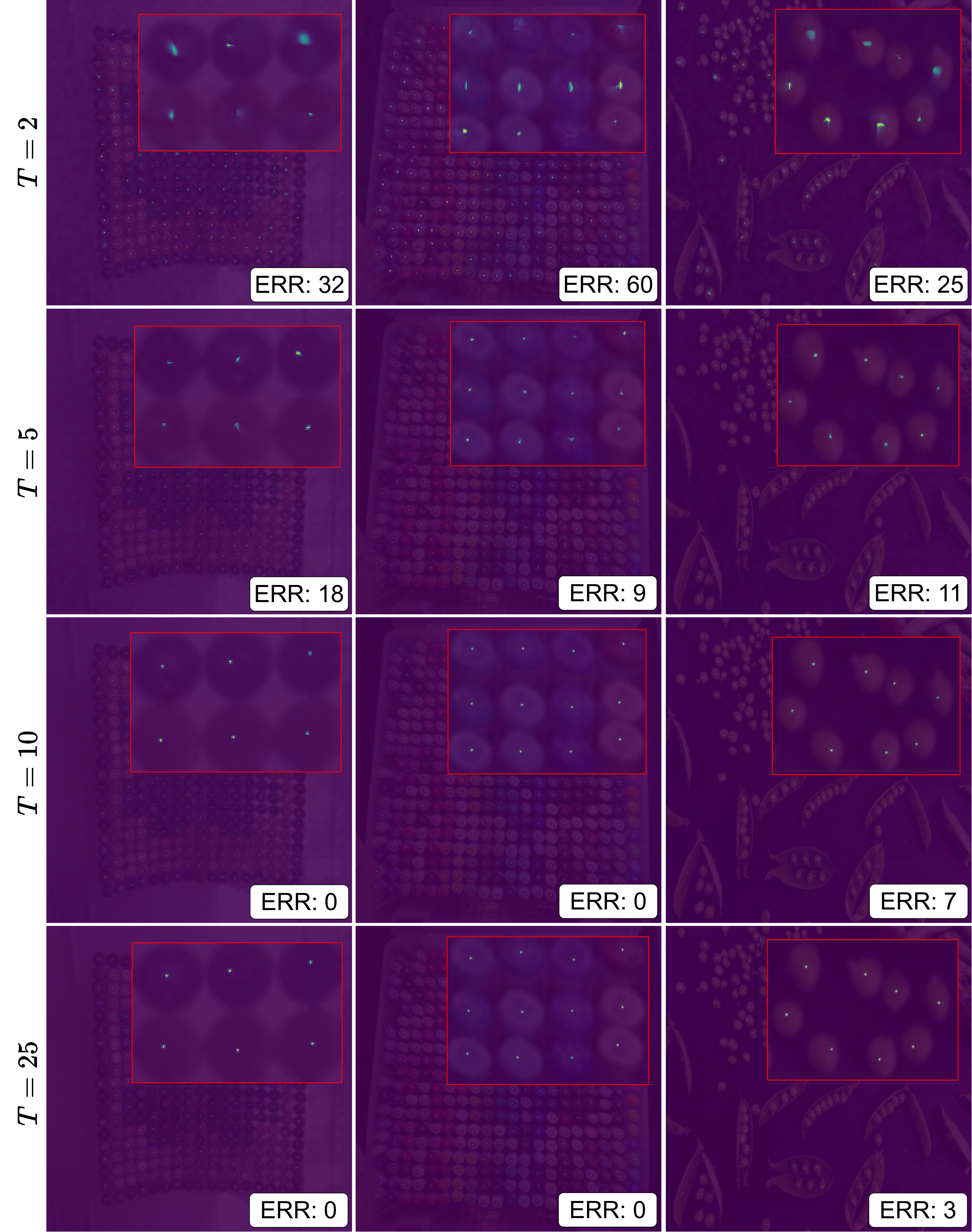}
    \caption{Reducing the diffusion process to $T=2$ timesteps, yields inadequate density maps for peak counting (numbers indicate the count error after applying NMS). Increasing to $T=25$ timesteps, \methodName{} converges, producing high quality peaks.}
    \label{fig:timesteps_quali}
\end{figure}

Furthermore, the impact of shape embeddings is evaluated by removing $\mathbf{p}^s$ from the prototypes in exemplar-based conditioning. 
\methodName{}$_{\overline{\mathrm{p}^s}}$ achieves 24\% drop in MAE and 32\% drop in RMSE, demonstrating the importance of exemplar shape information for few-shot counting. 
This observation is in line with findings in~\cite{djukic2023low}.

The importance of the AM-RADIO backbone is evaluated by replacing it with a VAE encoder~\cite{podell2023sdxl} pretrained for image reconstruction, 
a standard DINOv2~\cite{oquab2023dinov2} backbone and a ResNet~\cite{he2016deep} backbone pretrained on ImageNet~\cite{deng2009imagenet}. 
\methodName{}$_{\text{VAE}}$ achieves 30\% and 15\% performance drop in terms of MAE and RMSE, respectively. 
Such a substantial performance drop is attributed to the low-dimensional (4-channel) feature space of the VAE encoder. 
\methodName{}$_{\text{ResNet}}$ demonstrates a 26.9\% performance drop in MAE and a 11.8\% drop in RMSE compared to \methodName{}, showing comparable degradation in counting accuracy as \methodName{}$_{\text{VAE}}$.
\methodName{}$_{\text{DINOv2}}$ demonstrates a 15\% performance drop in MAE and 11\% drop in RMSE compared to the proposed \methodName{}, achieving comparable counting performance as the recent low-shot counter GeCo~\cite{pelhan2024novel}.
These results confirm that excellent performance of the proposed \methodName{} does not heavily depend on the selection of the image encoder.

To further verify that the AM-RADIO backbone alone does not account for \methodName{}'s gains,
we equip LOCA~\cite{djukic2023low}, with the same AM-RADIO backbone.
The suggested adjustment results in a 11.42 MAE / 40.01 RMSE on the FSC147
validation set and 10.84 MAE / 71.40 RMSE on the test set (from 10.24 / 32.56 and
10.79 / 56.97, respectively), indicating that the expressive features from AM-RADIO alone are not enough for state-of-the-art performance.
This supports our previous conclusion that \methodName{}'s improvement stems from the combination of strong features, latent diffusion over location maps, and adaptive exemplar conditioning, rather than the backbone alone.

Next, we evaluate the impact of using the standard test-time normalization.
Results in Table~\ref{tab:ablation_fsc147} show that turning off TTN (\methodName{}$_{\overline{TTN}}$) leads to a slight performance drop of 1.8\% RMSE, while maintaining a similar MAE to the baseline.
This showcases \methodName's robust ability to localize each object only once, \ie~\methodName{} avoids multiple detections of the same object.

\begin{table}[t]
\centering
\caption{Few-shot counters on the FSC147~\cite{ranjan2021learning} without upscaling and tiling~\cite{pelhan2026generalized}. Performance drops are marked in red.}
\label{tab:fscd147_noupscale}
\begin{tabular}{lcc}
\toprule
Method & MAE ($\downarrow$) & RMSE ($\downarrow$) \\
\midrule
PSECO   & 13.05 & 112.86 \\
LOCA    & 10.79 & 56.97  \\
DAVE    & 12.21\,\textcolor{red}{$\downarrow$1.8} & 94.93\,\textcolor{red}{$\downarrow$20.4} \\
GeCo    & 8.88\,\textcolor{red}{$\downarrow$1.0}  & 75.67\,\textcolor{red}{$\downarrow$21.4} \\
CountGD & 9.44\,\textcolor{red}{$\downarrow$1.0} & 98.84\,\textcolor{red}{$\downarrow$9.0}\\
CoDi & 9.53\,\textcolor{red}{$\downarrow$1.3} & 49.14\,\textcolor{red}{$\downarrow$13.4} \\
\bottomrule
\end{tabular}
\end{table}


We further show the importance of using diffusion during training and inference by training \methodName{}$_{\text{direct}}$, which replaces the iterative generation of location maps, with a direct regression of latent location maps.
\methodName{}$_{\text{direct}}$ demonstrates a substantial performance drop of 87.8\% MAE and 22.7\% RMSE. 
The results confirm that iteratively denoising latent location maps and, at inference, generating a single sharp sample from the learned target distribution is crucial for accurate count estimation via non-maxima suppression.

Additionally we analyze the impact of upsampling to higher resolutions before passing them to \methodName{} and compare the results to other state-of-the-art methods without their respective upsampling/tiling mechanisms. Table~\ref{tab:fscd147_noupscale} shows that \methodName{}'s performance drops similarly to other methods, resulting in an increase of 1.3 MAE and 13.4 RMSE on the FSC147 validation set.

The impact of the Gaussian kernel size $\sigma$ used to construct the training-time locations maps is analyzed in Table~\ref{tab:sigmas}.
Increasing or decreasing values from $\sigma=0.5$ leads to increased counts errors. 
We believe that large kernels lead to overlaps in dense regions and density over-smoothing, preventing accurate peak detection.
On the other hand, while small $\sigma$ values are ideal for dense object detection, they are difficult to accurately encode and decode by the pretrained VAE, as well as hard to reconstruct by the diffusion model. 
Based on our ablations we therefore select $\sigma$ = 0.5, which provides a suitable tradeoff between the two effects.

\begin{table}
    \centering
    \caption{
    Counting performance at different $\sigma$ values on the FSC147~\cite{ranjan2021learning} validation set.
    }
    \label{tab:sigmas}
    \resizebox{0.9\linewidth}{!}{
    \begin{tabular}{cccccc}
        \toprule
        
        $\sigma$ & 0.0 & 0.25 & \textbf{0.5} & 1.0 & 1.5 \\
         
        \midrule

         MAE (val) & 10.57 & 9.38 & \textbf{8.21} & 9.25 & 10.35 \\
         RMSE (val) & 45.59 & 39.8 & \textbf{35.67} & 41.09 & 46.27 \\
        
        \bottomrule
    \end{tabular}
    }
\end{table}

\begin{table}
    \centering
    \caption{
    Ablation study on $\tau_\text{max}$ and $d_{min}$ hyperparameters for NMS step on the FSC147 validation set. Bold indicates the baseline configuration.
    }
    \label{tab:ablation_hyperparams}
    \resizebox{\linewidth}{!}{
    \begin{tabular}{lcccccccc}
        \toprule
        \multicolumn{8}{c}{$\tau_\text{max}$} \\
        \midrule
        Value & 0.05 & \textbf{0.1} & 0.2 & 0.3 & 0.4 & 0.5 & 0.6 \\
        \midrule
        MAE (val)  & 8.24 & \textbf{8.21} & 8.21 & 8.22 & 8.27 & 8.92 & 11.18 \\
        RMSE (val) & 35.41 & \textbf{35.67} & 36.00 & 36.34 & 37.23 & 41.08 & 53.73 \\
        \midrule[0.4pt]
        \midrule[0.4pt]
        \multicolumn{8}{c}{$d_{min}$} \\
        \midrule
        Value & \textbf{3$\times$3} & 5$\times$5 & 7$\times$7 & 9$\times$9 & $11\times11$ & $13\times13$ & 15$\times$15 \\
        \midrule
        MAE (val)  & \textbf{8.21} & 8.22 & 8.22 & 8.24 & 8.26 & 8.30 & 8.42 \\
        RMSE (val) & \textbf{35.67} & 35.89 & 36.12 & 36.51 & 36.91 & 37.55 & 38.43 \\
        \bottomrule
    \end{tabular}
    }
\end{table}

Next, in Table~\ref{tab:ablation_hyperparams}, we ablate the hyperparameters of the NMS step we perform to count the number of predicted peaks in the final location map.
$\tau_\text{max}$ defines the threshold above which a peak is considered valid and
$d_{min}$ is the minimal region size around which no other peak can be detected.
As we can see, increasing $\tau_\text{max}$ yields worse results, since we gradually remove more and more valid peaks that were reconstructed with lower peak values. 
However, even when increasing $\tau_\text{max}$ from 0.1 to 0.4 the performance drop in MAE is only 0.7\%, showcasing \methodName{}'s remarkable robustness to this hyperparameter. 
Similarly increasing the $d_{min}$ results in slightly worse performance, because we impose a restriction that a region of size N$\times$N can only contain a single object, which is not always true, when $d_{min}$ is large. 
Despite this, the counting performance remains stable even when increasing $d_{min}$ from three to nine, with MAE dropping by only 0.3\%.

\begin{figure}
    \centering
    \includegraphics[width=\linewidth]{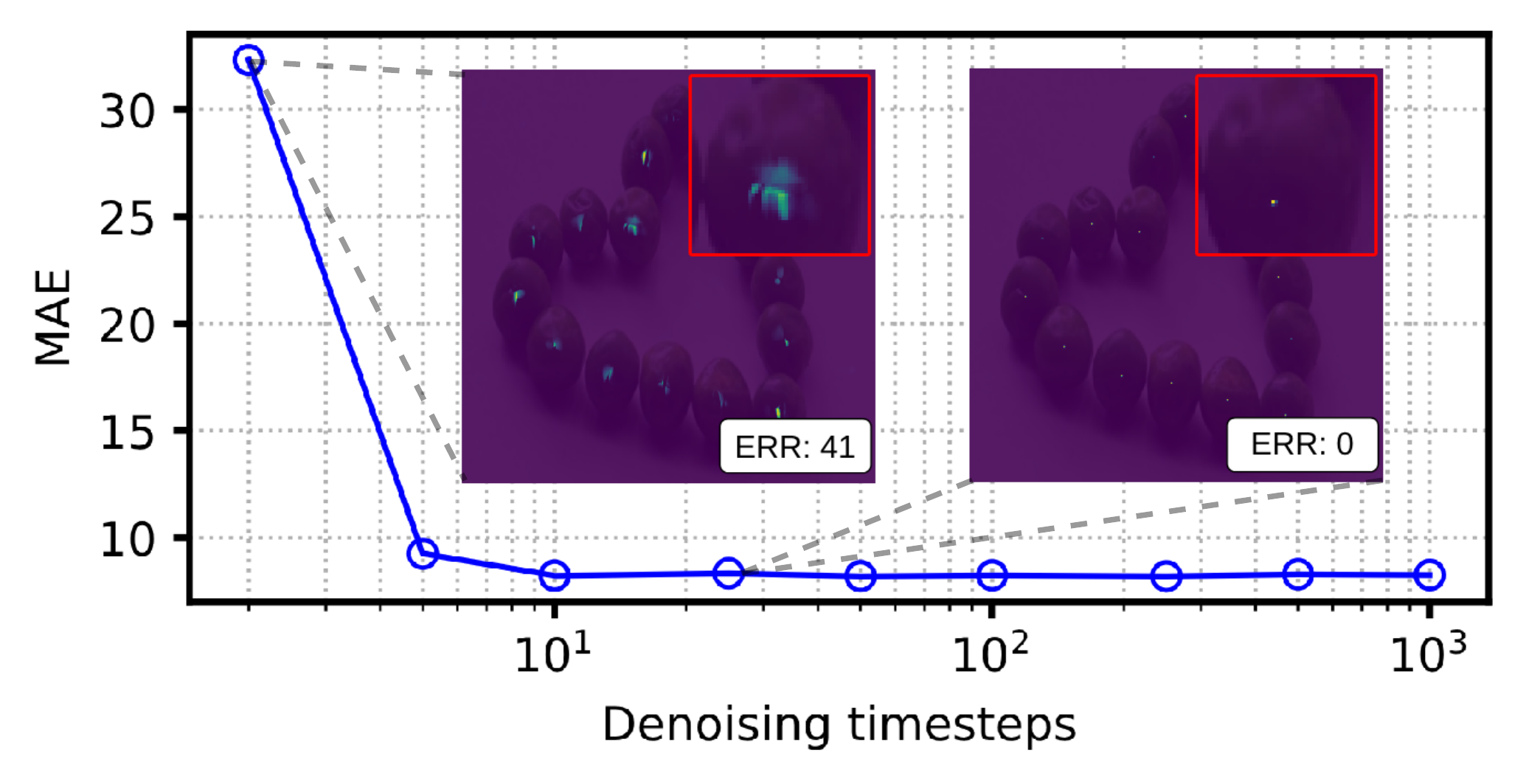}
    \caption{
        MAE on FSC147~\cite{ranjan2021learning} w.r.t. denoising timesteps, with two density maps visualized for an image at $T=2$ and $T=25$ timesteps, at which \methodName{} already converges.
    }
    \label{fig:timestep_ablation}
\end{figure}

Lastly, Figure \ref{fig:timestep_ablation} shows counting performance while varying the number of timesteps $T$ during inference. 
\methodName{} shows impressive robustness to the wide range of timestep number. 
The predicted Gaussian modes collapse to points and counts converge already at $T=10$ timesteps. 
Detailed qualitative comparisons at different timesteps are visualized in Figure \ref{fig:timesteps_quali}.

\subsection{Performance by Object Density}

\begin{figure}
    \centering
    \includegraphics[width=\linewidth]{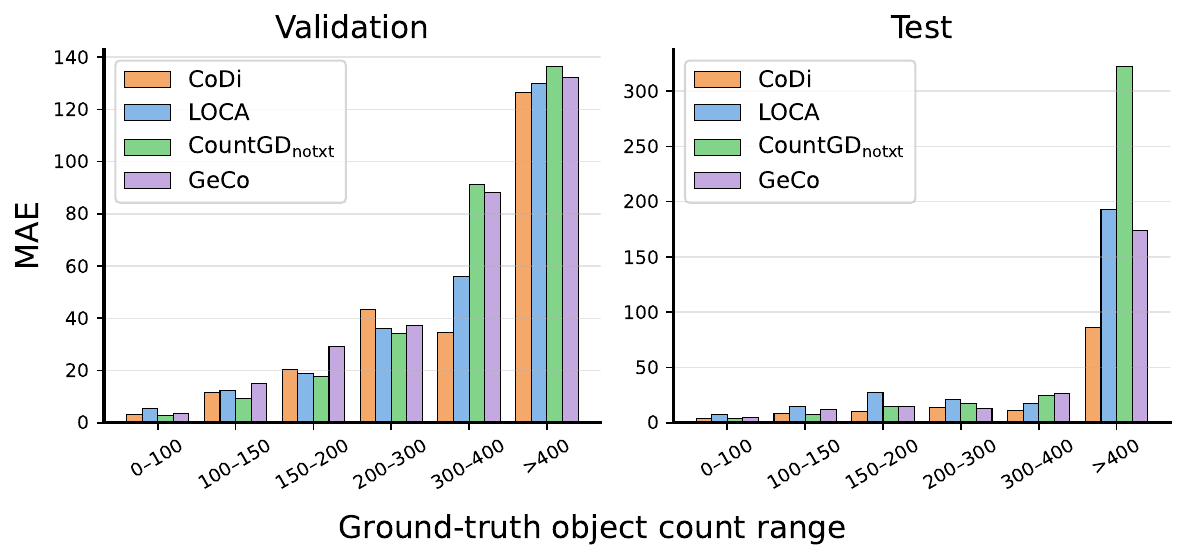}
    \caption{
        MAE on the FSC-147 validation (left) and test (right) splits, broken down by ground-truth object count range. On both validation and test sets \methodName{} maintains notably lower error than the other methods in the highest count range, indicating better robustness to dense scenes.
    }
    \label{fig:countbin_comparison}
\end{figure}

We analyze how performance varies with the number of objects present in an image, since dense scenes are known to be particularly challenging for counting methods.
Figure~\ref{fig:countbin_comparison} shows the MAE on the FSC147 validation and test splits, stratified by ground-truth object count. For images with fewer than 300 objects, all methods achieve broadly similar error levels, however when the number of objects in an image exceeds 300, \methodName{} shows notably better scaling, achieving substantially lower MAE. This suggests that \methodName{} is more robust to the density-related failure modes that affect competing methods. We argue that, the robustness arises from the reformulation of counting as a generation of high-fidelity location maps that contain clearly separable narrow peaks, that are easily countable by NMS.

\subsection{Robustness of counting performance}
\textbf{Repeatability.}
Since \methodName{} is built on a diffusion framework, the outputs are stochastic. To characterize the stability of the predictions, we thus calculated the statistics of the performance measures on the FSC147~\cite{ranjan2021learning} dataset over twenty different inference runs. Table~\ref{tab:stats_diff_model} shows remarkable stability with MAE standard deviation $\sigma_\text{MAE}=0.07$ (0.8\% relative deviation to the mean) and $\sigma_\text{RMSE}=0.38$ (1\% relative deviation to the mean). Additionally, we compute the 95\% confidence interval for MAE and RMSE and find that they are [8.24, 8.31] and [35.94, 36.30] respectively.
This means that when retraining our model from scratch the validation results are 95\% likely to fall within this range of values.
This confirms excellent repeatability and stability of \methodName{}.

\begin{table}
    \centering
    \caption{
    Mean ($\mu$) and std. deviation ($\sigma$) of counting performance measures on 20 inference runs on FSC147~\cite{ranjan2021learning} validation set.
    }
    \label{tab:stats_diff_model}
   \resizebox{0.6\linewidth}{!}
   {
    \begin{tabular}{cccc}
        \toprule
        $\mu_\text{MAE}$ & $\sigma_\text{MAE}$ & $\mu_\text{RMSE}$ & $\sigma_\text{RMSE}$ \\
        \midrule
        8.28 & 0.07 & 36.12 & 0.38 \\
        \bottomrule
    \end{tabular}
    }
\end{table}

        
        
\textbf{Exemplar quality.}
\begin{table}[ht!]
    \centering
    \caption{Re-run of \methodName{} with noisy exemplars at increasing bounding-box perturbation levels (Perturb.).}
    \resizebox{\linewidth}{!}{
    \begin{tabular}{llllll}
        \toprule
        & & \multicolumn{2}{c}{Validation set} \\
        \cmidrule(lr){3-4} \cmidrule(lr){5-6}
        Method & Perturb. & MAE ($\downarrow$) & RMSE ($\downarrow$) \\
        \midrule
        \multirow{3}{*}{CountGD$_{\text{notxt}}$}
            & 0\%  & 8.12  & 38.97\\ 
            & 5\%  & 8.19  & 39.25\\ 
            & 10\% & 8.21  & 39.36\\ 
        \midrule
        \multirow{3}{*}{\methodName{}}
            & 0\%  & 8.21  & 35.67\\ 
            & 5\%  & 8.24  & 35.52\\ 
            & 10\% & 8.29  & 35.62\\ 
        \bottomrule
    \end{tabular}
    }
    \label{tab:noisyexemplars}
\end{table}
For practical applications ideal exemplar bounding boxes are not always guaranteed, thus it is necessary for counters to remain robust to noisy exemplars. 
By adding noise scaled to 5\% or 10\% of the box size, to the exemplar box corners we can effectively simulate inaccuracies in the annotation process. 
We re-ran \methodName{} and CountGD$_{\text{notxt}}$~\cite{amini2024countgd} with the described setup and present the results in Table~\ref{tab:noisyexemplars}.
Both \methodName{} and CountGD$_{\text{notxt}}$ show remarkable robustness to exemplar annotation noise. 
With a 10\% perturbation \methodName{}'s performance drops by only 0.08 MAE and retains a similar RMSE as the baseline. CountGD$_{\text{notxt}}$ shows a similar trend, with a performance drop of 0.09 MAE and 0.39 RMSE on the FSC147 validation set.

\begin{figure*}
    \centering
    \includegraphics[width=0.93\linewidth]{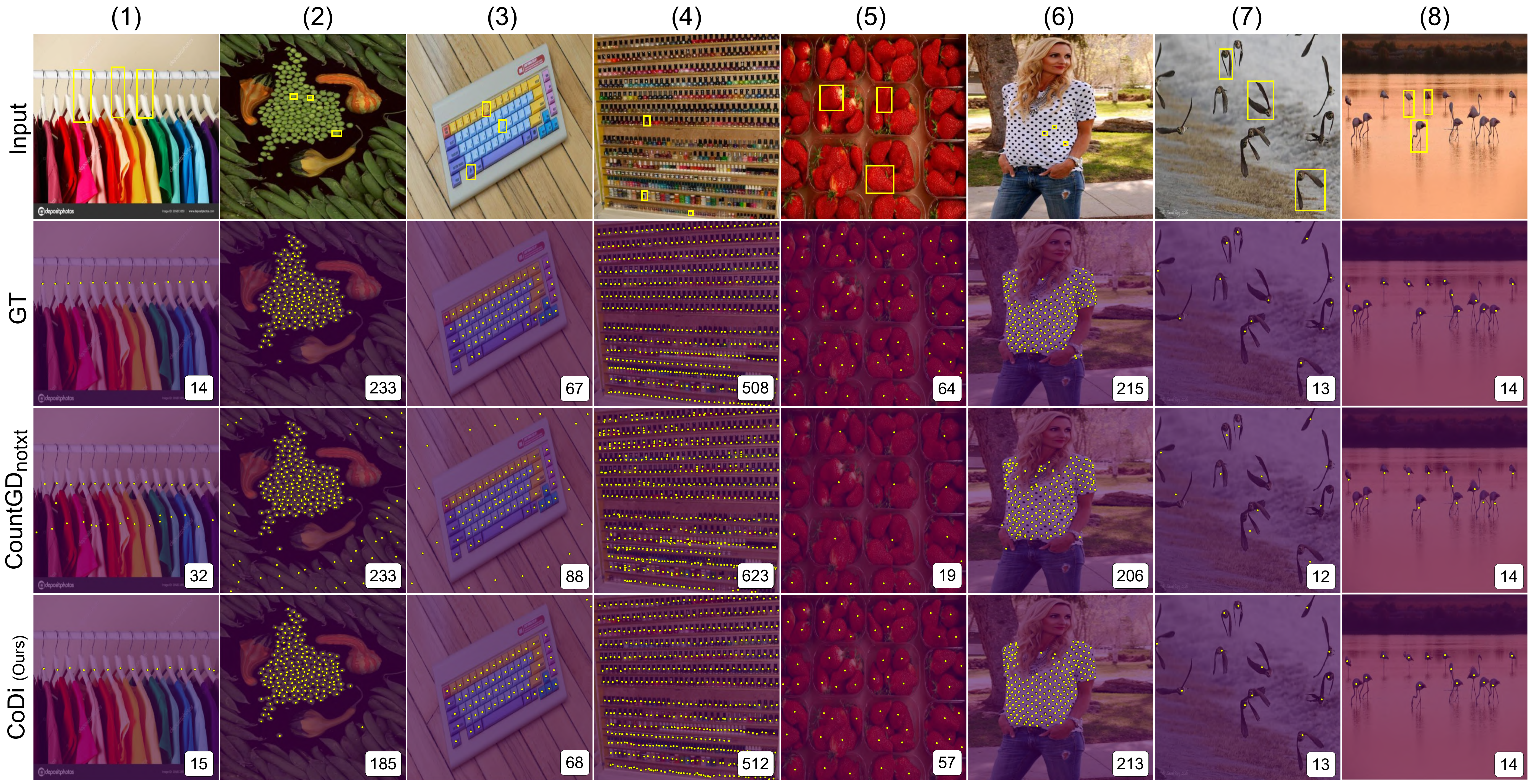}
    \caption{Qualitative comparison of \methodName{} with the current state-of-the-art CountGD$_{\text{notxt}}$~ \cite{amini2024countgd}. Input images with exemplars and ground truth are shown in the first two rows, while the rest show the detected object locations and their estimated total counts. 
    }
    \label{fig:qualitative}
\end{figure*}

\subsection{Counting by applying NMS on a predicted density map}

\begin{figure}
    \centering
    \includegraphics[width=\linewidth]{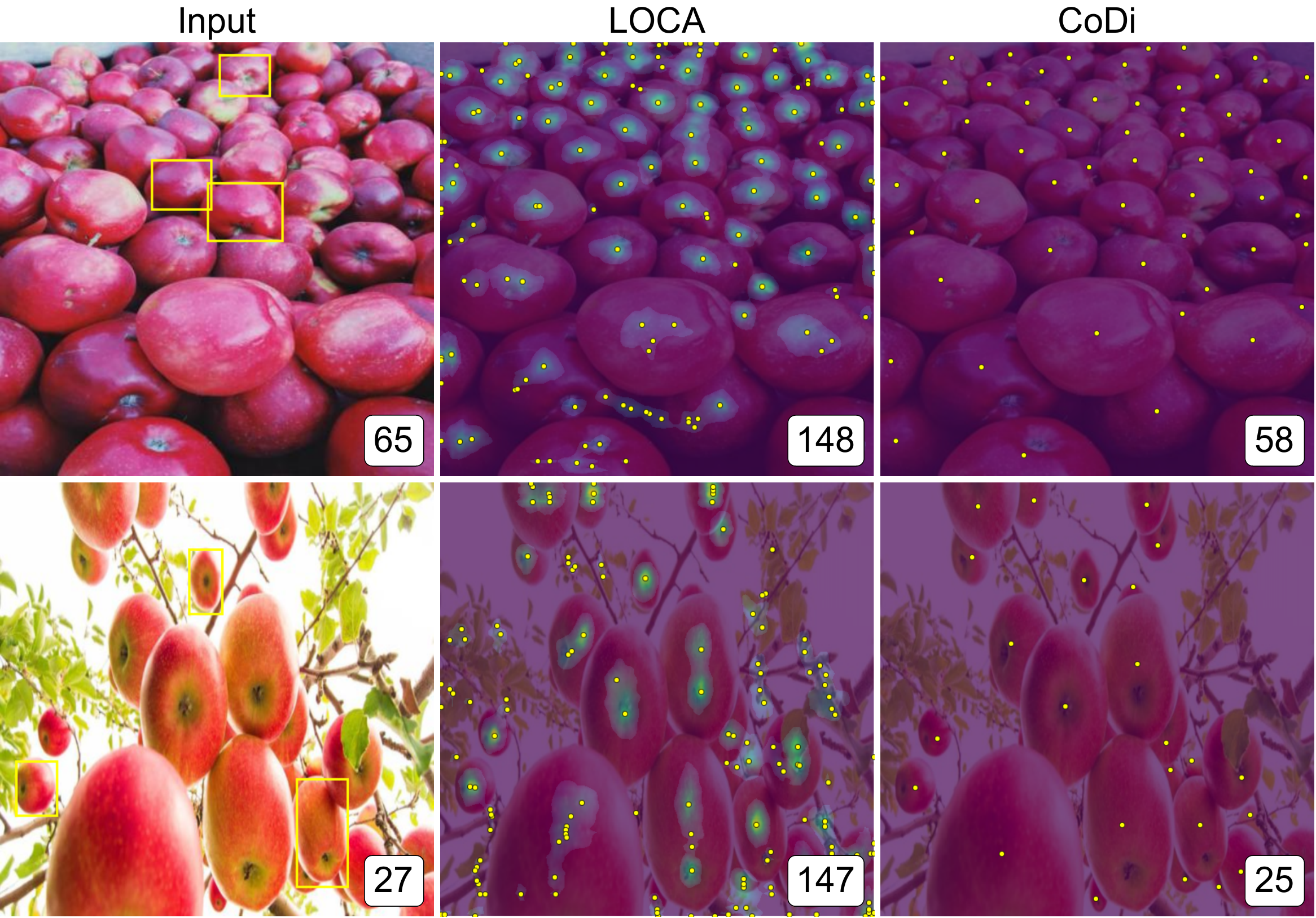}
    \caption{
        NMS applied on density-based few-shot LOCA~\cite{djukic2023low} compared to CoDi.
    }
    \label{fig:supp_img1}
\end{figure}

In this experiment we evaluate the ability of a density-based few-shot counter to predict object locations (centers) from predicted density maps. 
In particular, we use LOCA~\cite{djukic2023low} to get density map, apply a non-maxima suppression (NMS) and take the remained local maximum as object detections. 
The qualitative comparison of this LOCA version and the proposed \methodName{} is shown in Figure~\ref{fig:supp_img1}. 
LOCA predicts density maps with wide blob-like unit Gaussians which yield inaccurate peak locations and counts when processed with NMS. 
\methodName{}, on the other hand, produces narrow expressive peaks that are not ambiguous to count.
These examples also demonstrate the drawbacks of density-based methods when applied to scenes where the objects vary in sizes. 
The first row of Figure~\ref{fig:supp_img1} shows that LOCA produces strong responses to objects that coincide with the exemplar sizes, but weak and inaccurate responses on bigger objects closer to the camera. 
On the other hand, \methodName{} predicts the same size Gaussian kernels for all objects, yielding superior count estimations.

\subsection{Qualitative analysis}  \label{sec:qualitative}

\textbf{Qualitative comparison with CountGD$_{\text{notxt}}$.}
For a fair qualitative evaluation (see Figure \ref{fig:qualitative}) we compare \methodName{} to CountGD$_{\text{notxt}}$~\cite{amini2024countgd}, which is the no-text variant of the current state-of-the-art point-based counter CountGD~\cite{amini2024countgd}.
In columns 1,2,3 CountGD$_{\text{notxt}}$ does not specialize well to the exemplar appearance and detects also objects which are not specified as a target class, leading to a lot of false positive detections. 
In contrast, \methodName{} successfully specializes to the target object category with minimal (or no) false positives. 
A related case is shown in column 5 -- CountGD$_{\text{notxt}}$ completely ignores the exemplars (strawberry fruit) and counts whole boxes as objects. In contrast, \methodName{} adheres well to the exemplars due to diffusion-step exemplar specialization modules.
Column 4 shows a challenging case of dense regions, in which CountGD$_{\text{notxt}}$ results in many false detections, while \methodName{} delivers stable counts thanks to the gradual diffusion-based refinement process. 
Column 6 shows a challenging situation with dense and extremely small objects. CountGD$_{\text{notxt}}$ results in many missed detections, while \methodName{} comfortably detects their locations and successfully counts them.
Columns 7 and 8 show challenges with counting {\it non-blobby} objects. CountGD$_{\text{notxt}}$ results in inaccurate localization, while \methodName{} performs well due to the iterative process, which gradually refines positions of the individual detections conditioned on the adaptive prototypes.

\begin{figure*}
    \centering
    \includegraphics[width=0.9\linewidth]{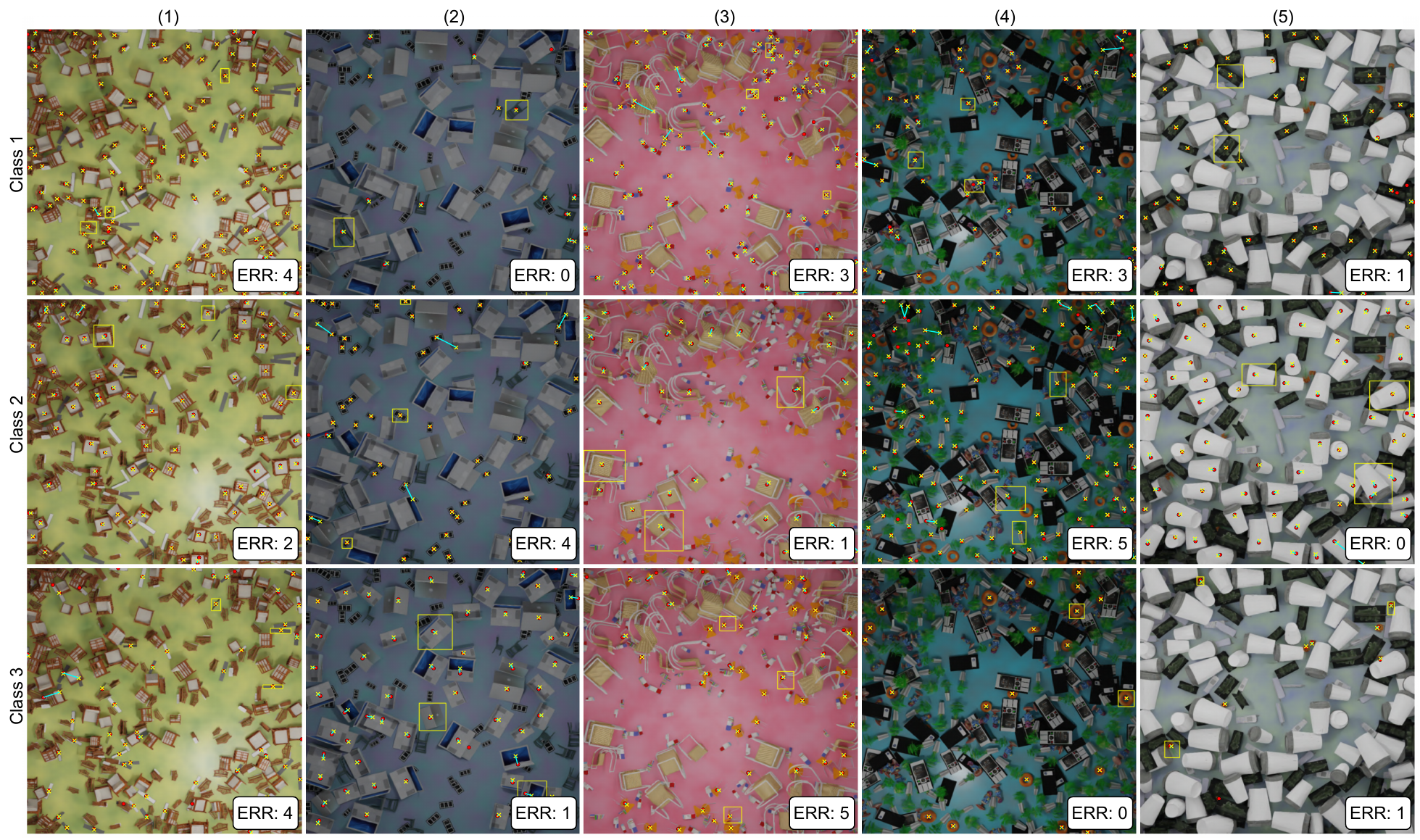}
    \caption{
        Qualitative results of \methodName{} on the MCAC dataset~\cite{hobley2025abc}. 
        Each column shows the same image, while different target classes are selected in each row. 
        Yellow crosses and red dots denote predicted and ground truth locations, respectively. 
        Large localization errors are denoted by a cyan line, which connects predicted location with the closest ground truth location.
        Counting errors are shown in the bottom right corners.
    }
    \label{fig:mcac_quali}
\end{figure*}

\textbf{Qualitative results on MCAC.}
In Figure \ref{fig:mcac_quali} we show qualitative results of \methodName{} on the MCAC dataset \cite{hobley2025abc}.
Each column displays the same image, with each row selecting a different target class. 
\methodName{} shows excellent robustness to distractors/objects of different categories from the selected exemplars, even in cluttered scenarios (column 1, 4, 5),
indicating that the proposed iterative exemplar conditioning mechanism is guiding the diffusion process to the correct solutions.
Additionally \methodName{} performs exceptionally well in situations where distractors and target category are visually similar, as can be seen in column 1 (rows 1 and 2), where both object categories have a similar color, yet \methodName{} is capable of correctly differentiating between them. 
Furthermore, column 3 (rows 2 and 3), shows that \methodName{} correctly discriminates even between objects of the same concept (\ie~chairs). 
Lastly, in column 5 (row 3), we can see that \methodName{} accurately detects even sparsely distributed small objects in a cluttered scene. 

\begin{figure*}[ht!]
        \centering
        \includegraphics[width=\linewidth]{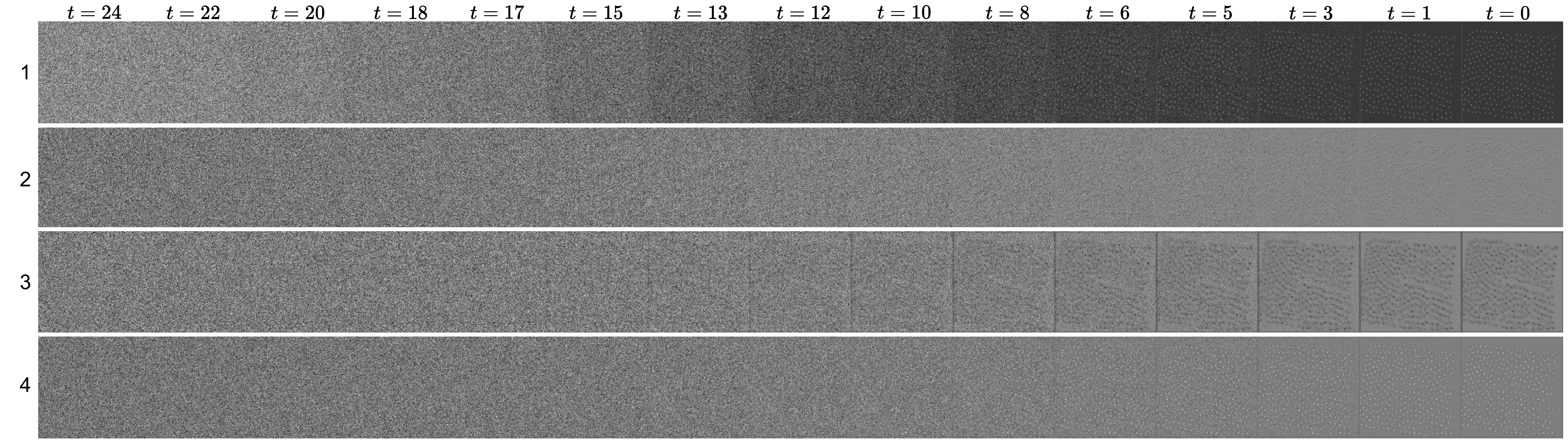}
        \caption{
            Generation of latent object location maps in \methodName{} with four latent VAE channels (rows) and 25 timesteps (columns).
        }
        \label{fig:vae_diff}
\end{figure*}

\textbf{Visualization of the latent diffusion process.}
To better understand the generation process of \methodName{}, we visualize the reverse diffusion process for each latent VAE channel in Figure~\ref{fig:vae_diff}.
\methodName{} showcases a stable denoising process, that gradually refines latent object location maps, with approximately equal changes in noise levels between timesteps.
While these changes are mostly imperceptible in the first half of the denoising process, we can clearly observe the gradual improvement in quality in the latter half of the process.

\begin{figure}
        \centering
        \includegraphics[width=\linewidth]{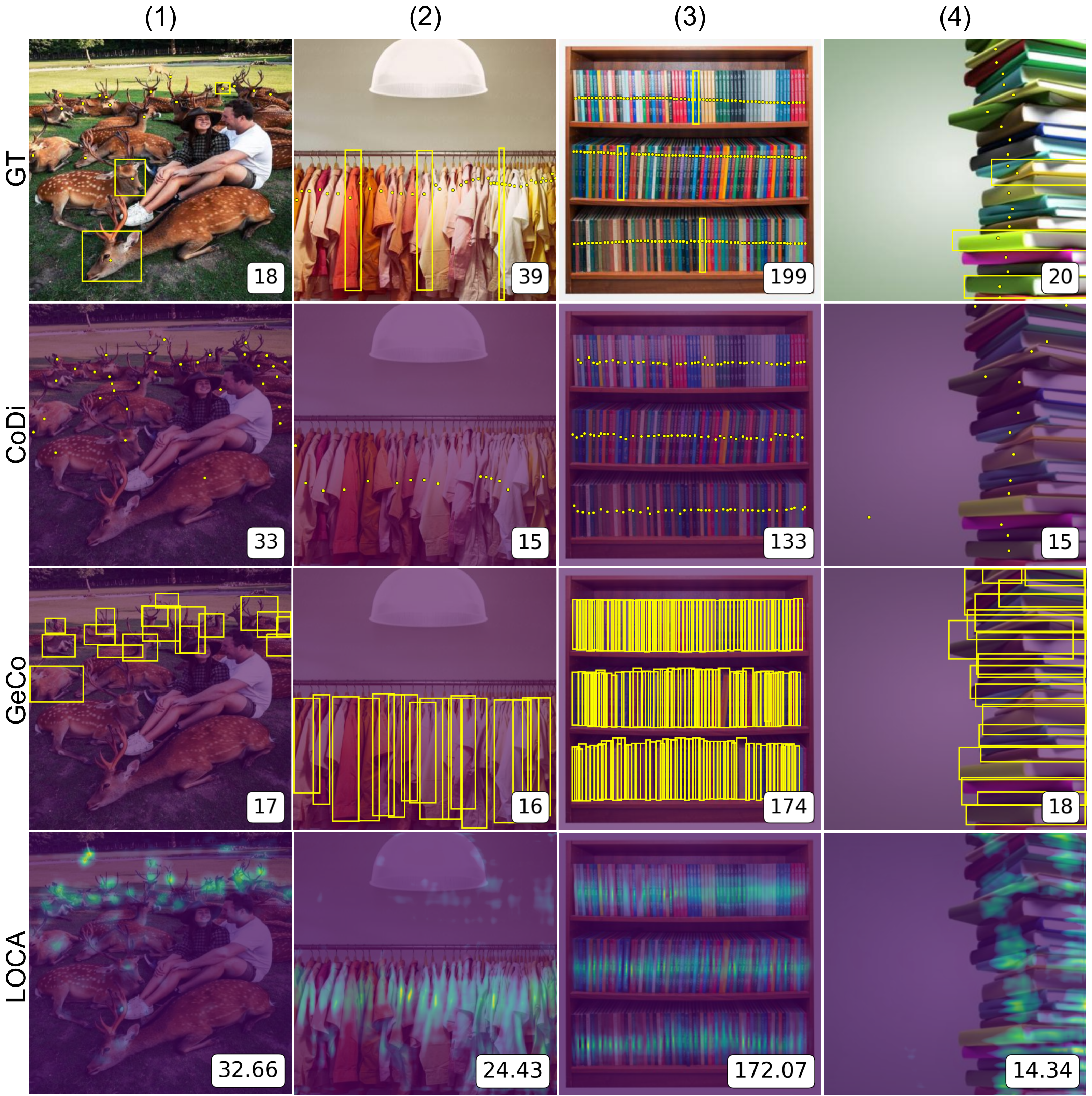}
        \caption{
            Failure cases of \methodName{} compared to GeCo~\cite{pelhan2024novel} and LOCA~\cite{djukic2023low}. First row denotes input images with exemplars (bounding boxes), ground-truth object centers and the object count. The remaining rows show method outputs and their estimated count.
        }
        \label{fig:supp_img2}
\end{figure}

\textbf{Failure cases.}
In Figure~\ref{fig:supp_img2} we visualize failure cases and compare the outputs to GeCo~\cite{pelhan2024novel} and LOCA~\cite{djukic2023low}. 
In column 1, \methodName{} fails to adapt to the exemplar appearance, which denotes only the deer's head and detects the whole animal. 
GeCo similarly places the predicted bounding boxes over the whole deer and even misses the ones in the foreground, while LOCA similarly to \methodName{}, overshoots the estimated count and yields an inaccurate density map.
Columns 2, 3 and 4 show that \methodName{} has difficulty with detecting objects with extreme aspect ratios when they are placed close together, or even overlapping (column 2).
In comparison, GeCo better detects and estimates the target counts of books in columns 3 and 4, while failing in a similar fashion as \methodName{} in column 2, displaying its ability to handle objects with extreme aspect ratio. 
Similarly, LOCA outperforms \methodName{} in columns 2 and 3, showing superior performance on densely packed objects, however it struggles to accurately estimate the target count in column 4, yielding a similar result to \methodName{}.

\section{Conclusion}
\label{sec:conclusion}

We proposed \methodName{}, a novel low-shot diffusion-based object counter that estimates object locations map in a latent space, and covers all low-shot scenarios. \methodName{} features a novel timestep-dependent exemplar conditioning mechanism tailored specifically for one- and few-shot counting scenarios, which improves object discrimination and counting performance, and can be easily adapted to reference-less setup.

On two challenging standard benchmarks \methodName{} surpasses the best counters by approximately 15\% in total count MAE, 
achieving state-of-the-art performance in a few-shot counting setup, and shows a remarkable robustness to the number of provided exemplars. 
\methodName{} also outperforms the best reference-less and stays competitive with recent zero-shot methods, thus setting a new state-of-the-art across the low-shot counting spectrum.

The location prediction capability is sufficient for most counting and localization applications. Nevertheless, in our future work, we plan to explore bounding box prediction, which will broaden the scope towards general object detection, as well as address some of the failures arising from non-blobby objects with extreme aspect ratios.

\section*{Acknowledgments}
This work was supported by Slovenian research agency program P2-0214 and projects J2-2506, L2-3169, Z2-4459 and J2-60054, and by supercomputing network SLING (ARNES, EuroHPC Vega - IZUM).

\bibliographystyle{IEEEtran}
\bibliography{egbib}

 



\begin{IEEEbiographynophoto}{Grega Šuštar} 
received the B.Sc. and M.Sc. degrees in computer and information science from the University of Ljubljana, Ljubljana, Slovenia, in 2022 and 2025 respectively.
He is currently a Researcher with Visual Cognitive Systems Laboratory (ViCoS), Faculty of Computer and Information Science, University
of Ljubljana. His research interests include object counting and detection, image-based geolocalization, computer vision, and deep learning.
\end{IEEEbiographynophoto}

\begin{IEEEbiographynophoto}{Jer Pelhan} 
received the B.Sc. and M.Sc. degrees in computer and information science from the University of Ljubljana, Ljubljana, Slovenia, in 2021 and 2023 respectively, where he is currently pursuing the Ph.D. degree with the Visual Cognitive Systems Laboratory (ViCoS), Faculty of Computer and Information Science.
He is currently a Researcher with ViCoS, Faculty of Computer and Information Science, University of Ljubljana. His research interests include object counting and detection, multi-target object tracking, computer vision, and deep learning.
\end{IEEEbiographynophoto}

\begin{IEEEbiographynophoto}{Alan Lukezic} 
received the Ph.D. degree from the Faculty of Computer and Information Science, University of Ljubljana, Slovenia in 2021. 
He is currently with the Visual Cognitive Systems Laboratory, Faculty of Computer and Information Science, University of Ljubljana, as a Teaching Assistant and a Researcher. 
His research interests include computer vision, data mining and machine learning.
\end{IEEEbiographynophoto}

\begin{IEEEbiographynophoto}{Matej Kristan} 
is a full professor and a vice chair of the department of artificial intelligence at the Faculty of Computer and Information Science, University of Ljubljana. He leads the Visual object tracking VOT initiative, serves as Associate Editor of IJCV and was president of the IAPR Slovenian pattern recognition society (2021-2025). He has co-organized over sixteen workshops and conferences, and received thirty research excellence and teaching awards, including ISPA2015 and BMVC2022, the 2024-23 best paper award at the Pattern Recognition journal. His research interests include visual object tracking, few-shot detection, perception methods for autonomous boats, anomaly detection, and machine-learning-based physics prediction models.
\end{IEEEbiographynophoto}

\vfill

\end{document}